\documentclass{article}

\usepackage[final,eandd]{neurips_2026}

\usepackage[utf8]{inputenc}
\usepackage[T1]{fontenc}
\usepackage{url}
\usepackage{xurl} %
\usepackage{booktabs}
\usepackage{float}
\usepackage{placeins}
\usepackage{amsmath}
\usepackage{amsfonts}
\usepackage{nicefrac}
\usepackage{microtype}
\usepackage{xcolor}
\definecolor{linkblue}{HTML}{3B82F6}
\usepackage{tabularx}
\usepackage{longtable}
\usepackage{array}
\usepackage{fontawesome5}
\usepackage[breaklinks=true]{hyperref}
\newcolumntype{L}[1]{>{\raggedright\arraybackslash}m{#1}}
\DeclareUnicodeCharacter{00A7}{\S}
\usepackage[most]{tcolorbox}
\tcbuselibrary{listings,breakable}
\usepackage{listings}

\lstdefinestyle{promptstyle}{
  basicstyle=\ttfamily\scriptsize,
  breaklines=true,
  breakatwhitespace=false,
  columns=fullflexible,
  keepspaces=true,
  showstringspaces=false,
  upquote=true,
  literate={§}{{\S}}1,
  xleftmargin=0pt,
}

\newcommand{\prompttcbinput}[4]{%
  \tcbinputlisting{
    listing engine=listings,
    listing file={#1},
    listing only,
    listing options={style=promptstyle},
    breakable,
    enhanced,
    colback=gray!3,
    colframe=gray!25,
    coltitle=white,
    colbacktitle=black!75,
    colframe=black!75,
    boxrule=1pt,
    arc=2pt,
    outer arc=2pt,
    left=1.8em,
    right=1.8em,
    top=1.2em,
    bottom=1.2em,
    boxsep=0pt,
    titlerule=0pt,
    toptitle=0.45em,
    bottomtitle=0.45em,
    lefttitle=1.2em,
    righttitle=1.2em,
    fonttitle=\bfseries\large,
    title={#2},
    label={lst:#4}
  }%
}

\title{\textsc{JudgmentBench}: Comparing Rubric and Preference Evaluation for Quality Assessment}

\author{%
\normalfont
\begin{tabular}{c}
\textbf{Russell~Yang}\textsuperscript{1}\quad
\textbf{Ruishi~Chen}\textsuperscript{1}\quad
\textbf{Pierce~Kelaita}\textsuperscript{1}\quad
\textbf{Riya~Ranjan}\textsuperscript{1}\quad
\textbf{Sibo~Ma}\textsuperscript{1}\\[0.25em]
\textbf{Charles~Dickens}\textsuperscript{2}\quad
\textbf{Matthew~Guillod}\textsuperscript{3}\\[0.25em]
\textbf{Megan~Ma}\textsuperscript{1}\quad
\textbf{Julian~Nyarko}\textsuperscript{1}\\[0.3em]
{\footnotesize \textsuperscript{1}Stanford University\quad
\textsuperscript{2}Snorkel AI\quad
\textsuperscript{3}Harvey}\\[0.3em]
{\footnotesize
\href{mailto:jnyarko@law.stanford.edu}{\texttt{jnyarko@law.stanford.edu}}}\\[0.1em]
{\footnotesize \textit{Dataset:}
\href{https://huggingface.co/datasets/judgmentbench/JudgmentBench}{\textcolor{linkblue}{\texttt{huggingface.co/datasets/judgmentbench/JudgmentBench}}}}\\[0.1em]
{\footnotesize \textit{Code:}
\href{https://github.com/liftlab-SLS/JudgmentBench}
{\textcolor{linkblue}{\texttt{github.com/liftlab-SLS/JudgmentBench}}}}
\end{tabular}%
}

\begin{document}

\maketitle

\begin{abstract}
Two methodologies dominate current practices of benchmarking: rubric-based scoring evaluates items against predefined criteria, whereas comparative judgment elicits pairwise preferences between outputs. Although both methodologies are widely used, the choice between them is rarely justified. We release \textsc{JudgmentBench}, a benchmark of 30 real-world legal tasks, paired with 1,539 rubric scores and 1,530 pairwise preference judgments collected from practicing attorneys--including at major U.S. law firms--with substantial experience. The annotations constitute the first publicly available dataset in a high-expertise domain in which both supervision signals are elicited from the same pool of experts on the same tasks. Using LLM-generated outputs at three constructed quality levels, we provide an initial empirical comparison: comparative judgments recover the intended quality ordering substantially better than rubrics under both a per-task rank-correlation metric (mean Spearman's rank correlation of 0.908 vs. 0.150, $\widehat{D}_{\rho}=0.758$ [0.494, 1.021]) and a per-judgment pairwise win-rate metric (0.669 vs. 0.542, $\widehat{D}_{W}=0.127$ [0.067, 0.186]), while requiring less than half the annotation time. The patterns hold for human annotators and LLM autograders. Beyond this initial comparison, the paired structure of the dataset supports a broader research agenda on how expert judgment should be elicited, aggregated, and used as supervision in domains without verifiable ground truth.
\end{abstract}

\section{Introduction}
Large language models (LLMs) are increasingly deployed in high-stakes, high-expertise domains like medicine \citep{singhal_large_2023,pierson_using_2025,omiye_large_2024}, law \citep{guha_legalbench_2023,fei_lawbench_2024}, and finance \citep{wu_bloomberggpt_2023,yang_fingpt_2023,xie_pixiu_2023}. Many of the underlying scenarios are particularly judgment rich, rendering the measurement of output quality of these models notoriously difficult. Practical approaches vary widely across investigators and contexts, with two methodologies dominating current evaluation efforts: Under rubric-based scoring, outputs are graded against predefined criteria \citep{arora2025healthbench,shi_plawbench_2026,sharma_researchrubrics_2025,brookhart_appropriate_2018,jonsson_use_2007}. In contrast, under comparative judgment (CJ), annotators compare two outputs and identify the one they prefer \citep{thurstone_law_1927,pollitt_method_2012,lesterhuis2017comparative,verhavert_meta-analysis_2019}. In the benchmarking literature, the choice between the two methodologies often appears arbitrary and is rarely justified, raising questions as to whether and when one approach might yield more reliable assessments over the other.

We introduce a new type of benchmark dataset designed to allow for direct comparisons of rubric-based scoring and comparative judgment as assessment methodologies. Our study is grounded in the legal domain, capturing realistic work processes in large-scale commercial legal practice. This setting is often characterized by high levels of complexity in which attorneys accumulate and rely on experience acquired over many years of practice. Defining ``quality'' in such a setting is particularly challenging as legal work products are frequently multidimensional, context-dependent and resist straightforward objective evaluation criteria~\citep{ma_conceptual_2023,scott_hoffman_2009}. For example, outcomes alone are a poor proxy for quality in summary judgment briefing. Strong briefs may fail due to adverse facts or judicial preferences, while weaker ones may prevail when the law clearly favors one side \citep{tippett_does_2022}. In contexts like these, the choice of assessment methodology is itself part of the problem of defining legal quality.

The significance of comparing rubric-based scoring with comparative judgment extends beyond benchmark design. As LLMs are applied to economically consequential knowledge work, expert judgment must be represented in forms that can support systematic methods (e.g., evaluation, quality assurance, model post-training) at scale. Rubric-based scoring and comparative judgment represent two different approaches to this conversion. Rubrics are costly to create because they require experts to specify the dimensions of quality in advance, but once created, it is often assumed they are easy to automate and scale across systems \citep{starace2025paperbench,liu_openrubrics_2025}. Comparative judgments, by contrast, may better preserve tacit and holistic dimensions of expertise, including nuance, creativity, strategic judgment, and persuasive force, especially where quality is difficult to enumerate in advance \citep{jonsson_use_2007,pollitt_method_2012}. Yet because pairwise preference judgments are typically aggregated into rankings or latent utility estimates, the corresponding evaluations are harder to decompose into reusable criteria or audit after aggregation. This tradeoff matters not only for how AI systems are evaluated, but also for how expert judgment is incorporated into downstream systems such as model post-training.

Against this backdrop, our contributions are threefold:

\begin{enumerate}
    \item \textbf{A benchmark dataset of real-world tasks.} We release a set of high-complexity tasks drawn from Harvey's BigLaw Bench, selected to be representative of economically valuable legal work rather than stylized exam-style problems.
    \item \textbf{Paired rubric and preference annotations from legal experts.} Each task is annotated by experienced, practicing lawyers along both dimensions--atomistic, expert-developed rubric criteria and pairwise preference judgments--representing approximately \$48{,}500 in equivalent attorney time.\footnote{We value time contributed through partner firms at \$1,000 per hour and time contributed through Snorkel AI at \$125 per hour. This is an imputed value, not actual expenditure, as substantial contributions were pro bono.} To our knowledge, this is the first dataset in any high-expertise domain in which both supervision signals are elicited from the same pool of experts on the same tasks. This pairing enables comparisons within a shared task set and annotator pool.
    \item \textbf{A direct empirical comparison of rubric-based scoring and comparative judgment in law.} We find that comparative judgments are consistently better than rubrics for recovering the constructed quality ordering while requiring less than half the annotation time. The result holds for both human annotators and LLM autograders under our quality construction and the supplied rubrics, motivating further study of how evaluation and reward signals should be constructed in high-stakes professional domains.

\end{enumerate}

\section{Background and Related Work}
\label{sec:related_work}

\subsection{Conceptual Foundations}

Benchmarks are essential infrastructure for evaluating the quality of LLM outputs, providing standardized conditions under which models can be compared, progress tracked, and failure modes surfaced \citep{liang2022holistic,chang2024survey,qian2026benchmark,reuel2024betterbench}. The reliability of any benchmark, however, depends not only on the tasks it includes but on the methodology it uses to elicit and aggregate human or model judgments. These methodological choices long predate the era of LLM evaluation and have been studied extensively in the educational \citep{jonsson_use_2007,brookhart_appropriate_2018,pollitt_method_2012,verhavert_meta-analysis_2019} and psychological \citep{thurstone_law_1927,maydeu-olivares_structural_2005,stellmack_assessment_2009} measurement literature. Two strategies have historically dominated: rubrics, in which a response is scored against a set of predefined criteria, and comparative judgments, in which pairwise preferences are aggregated into an ordering.

Rubrics rest on the premise that quality can be made explicit by decomposing an artifact into components, scoring each separately, and aggregating the results. This decomposition gives rubrics their characteristic strengths: transparency, consistency, and a documented basis for each score that constrains rater discretion to predefined dimensions \citep{jonsson_use_2007,jonsson_transparency_2014,arora2025healthbench}. These properties make rubrics attractive in large-scale benchmarking, where they support standardized, administrable, and auditable evaluation across many tasks, models, and annotators \citep{he2025advancedif,hashemi-etal-2024-llm,starace2025paperbench,sharma_researchrubrics_2025,akyurek2025prbench}. The same decomposition, however, is the source of their documented limitations. The assumption that quality decomposes cleanly into a fixed set of criteria can break down for complex artifacts where the relevant dimensions are themselves contested or difficult to articulate in advance \citep{pollitt_method_2012,shankar_validators_2024,szymanski_designing_2026}. Rubric scoring is also susceptible to halo effects, in which a rater's overall impression of an artifact influences ratings on individual criteria \citep{ayoobiyan2023detecting,lai2015differentiation}. And because rubrics constrain attention to predetermined dimensions, they may systematically miss aspects of quality the rubric author did not anticipate, particularly the tacit, holistic features that experienced evaluators recognize but find difficult to enumerate \citep{shen_rethinking_2026,sadler_indeterminacy_2009,bloxham_mark_2011}. These limitations motivated a partial shift toward comparative judgments, building on Thurstone's observation that raters discriminate more reliably between two stimuli than they assign absolute values to a single stimulus in isolation \citep{thurstone_law_1927}. Subsequent work has shown that comparative judgments mitigate the unreliability of rubric-based grading for complex artifacts, such as essays \citep{pollitt_method_2012,verhavert_meta-analysis_2019,sims_rubric_2020}, impose lower cognitive demands on raters \citep{lesterhuis2017comparative}, can compare favorably on labor cost \citep{steedle_evaluating_2016}, and surface latent dimensions of quality that resist pre-specification, such as analytical depth, creativity, and persuasiveness \citep{kimbell2012evolving,kimbell2022examining,van2017complexity}.

Despite these well-characterized strengths and weaknesses, direct empirical comparisons of the two methodologies remain scarce, in part because no methodology-independent ground truth is available: each methodology is itself a definition of quality, and the construct cannot simultaneously serve as the standard against which competing definitions are judged. Prior work has approached this problem through three indirect strategies. The first compares methodologies through inter-rater reliability (IRR). Both rubrics and comparative judgments can achieve high IRR \citep{verhavert_meta-analysis_2019,jonsson_use_2007,sims_rubric_2020}, though the mechanisms differ: rubrics constrain rater attention and reduce the dimensionality of the judgment \citep{jonsson_use_2007}, while comparative judgments leverage the cognitive ease of relative discrimination \citep{verhavert_meta-analysis_2019}. High agreement under either mechanism, however, does not guarantee that raters are tracking the construct of interest \citep{moss_can_1994,bean2025measuring}; they may instead converge on surface cues only loosely correlated with quality, a concern echoed in NLP work on human evaluation and annotation artifacts \citep{huot_reliability_1990,crossley2011good,clark_all_2021,gururangan_annotation_2018} and inherited by LLM-based scorers, which exhibit sensitivity to verbosity, position, and other construct-irrelevant cues \citep{zheng_judging_2023,dubois_length_2024,chen_humans_2024,walsh2026measuring}.

A second strategy, \textbf{criterion validity}, uses external indicators as proxies for quality. When objectively verifiable answers exist, this can be done by pairing correct and incorrect responses and testing whether a judge recovers the ground-truth ordering \citep{tan_judgebench_2025}, or, in mathematics education, by correlating comparative judgments with conventional examination marks \citep{jones_problem_2015}. But such proxies are harder to construct in the legal domain. One might attempt to anchor evaluations to case outcomes--for instance, by asking annotators which of two briefs was filed by the prevailing party--but such designs are unlikely to isolate quality. Case outcomes are determined by many factors beyond the quality of the brief, including the strength of the underlying law and facts, judicial preferences, and strategic considerations outside the drafter's control. An evaluation methodology that tracked outcomes well in such a design might be tracking sensitivity to those confounders rather than the construct of interest, and one that tracked outcomes poorly might still be measuring quality accurately. A third strategy, \textbf{construction validity}, systematically manipulates output along known dimensions of quality and examines whether each methodology recovers the manipulation signal. Prior work has done so by varying model size and prompting strategy \citep{kim_aligning_2023}; related alignment work has used constructed preference data, including LLM-prompted preferences for direct preference optimization (DPO) training \citep{yuan_self-rewarding_2024} and rubric-generated synthetic preferences \citep{gallego_configurable_2025}. Our approach induces quality variation through prompting and aligns most closely with this third strategy. Most directly related, \citet{tripathi_pairwise_2025} compare pairwise and pointwise feedback protocols for bias in LLM-based evaluation; we extend the question to human expert evaluators and to a high-expertise professional domain, using paired annotations from the same pool of experts on the same tasks.

\subsection{Post-Training}

Beyond evaluation, rubric-based and preference-based feedback serve as the two dominant forms of supervision for post-training large language models in domains without verifiable rewards. Preference-based feedback, elicited as pairwise preference judgments and aggregated into a reward signal, underlies reinforcement learning from human feedback~\citep{ouyang_training_2022} and DPO~\citep{rafailov_direct_2023}, both of which are now standard tools for aligning model outputs with human judgment. More recently, rubric-based feedback has been used to construct rewards for reinforcement learning~\citep{huang2025rubricanchors}, with rubric-trained policies reported to outperform preference-trained reward models on instruction-following~\citep{viswanathan2025checklistsbetterrewardmodels} and direct or reference-based Likert judging in medicine and science~\citep{gunjal2025rubricsrewardsreinforcementlearning}.

A common assumption of much of this work is that rubrics generated from reference answers are roughly as informative as rubrics authored by domain experts~\citep{gunjal2025rubricsrewardsreinforcementlearning}. The extent to which this assumption holds in judgment-rich professional domains, where reference answers are themselves contested, is an open question. A parallel line of work has begun to bridge the two supervision signals: rubrics have been used to generate synthetic preference data~\citep{gallego_configurable_2025}, preference data have been used to learn rubric generators~\citep{lv_learning_2026,liu_openrubrics_2025}, and rubrics have been derived adaptively from pairwise comparisons during training \citep{rezaei2025onlinerubricselicitationpairwise}. Direct comparison of the two signals as supervision in expert domains, however, has been limited by data availability--specifically, by the absence of paired rubric and preference annotations from the same pool of experts on the same tasks. Our dataset is structured to make this comparison feasible.

\subsection{Legal Benchmarking}

In addition to the literature on evaluation and post-training, our work also contributes to an extant body of work on legal benchmarking. Currently, there is no settled standard for the choice of evaluation method in this space.

Early legal benchmarks largely evaluated models against predefined targets rather than open-ended expert assessments. LegalBench \citep{guha_legalbench_2023} compiled collaboratively curated legal reasoning tasks scored against task-specific labels or reference answers; LexGLUE \citep{chalkidis-etal-2022-lexglue} assembled fixed-label language understanding tasks; CaseHOLD \citep{zheng2021casehold} posed multiple-choice legal reasoning tasks; SARA \citep{holzenberger2020statutory} targeted statutory entailment and rule application; and LawBench \citep{fei_lawbench_2024} aggregated broader collections of legal reasoning evaluated against labels, answer keys, or reference outputs. Although some of these targets were expert-annotated, these evaluations treat recovering predefined answers as a proxy for quality. 

As the field has moved toward open-ended, real-world legal work products, evaluation has correspondingly shifted toward expert assessment rather than fixed-answer recovery. Many of the expert-assessment benchmarks in professional practice have been developed in industry settings and are not fully public. Some use rubrics, including Harvey's BigLaw Bench \citep{harvey_team_introducing_2024}, Thomson Reuters' CoCounsel benchmarking infrastructure \citep{thomson_reuters_scorecard_2025}, and Vals AI's legal AI evaluations \citep{vals_legal_ai_report_2025}. Others use comparative judgments, including GDPval \citep{patwardhan_gdpval_2025}, which spans professional tasks across domains including law, and Scale AI's SEAL leaderboards \citep{scale_ai_seal_2024}. Although data samples or small subsets are sometimes available, these benchmarks generally remain restricted.

Existing expert-assessment benchmarks for legal AI thus share two limitations: a lack of consistency in evaluation criteria across protocols, and the closed nature of those benchmarks that do collect expert labels. \textsc{JudgmentBench} addresses both. To our knowledge, it is the first publicly available expert-annotated legal benchmark in which outputs from the same tasks are evaluated under both rubric-based and comparative-judgment protocols, enabling direct comparison of the two methodologies on shared, high-judgment legal work products.

\section{Dataset Construction}
\label{sec:dataset-construction}
\begin{figure}[ht]
  \centering
  \includegraphics[width=\textwidth]{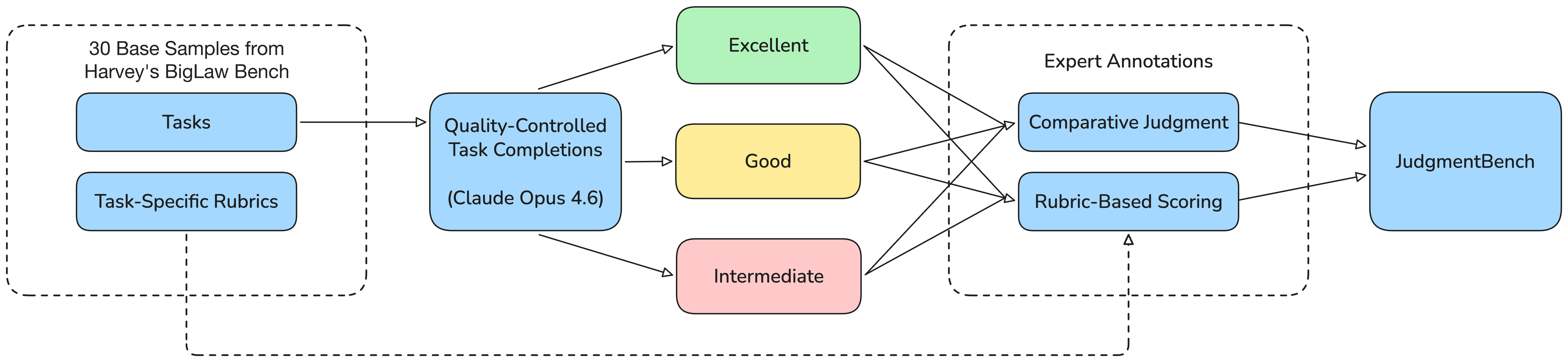}
  \caption{Illustration of dataset construction.}
  \label{fig:pipeline}
\end{figure}

Our full dataset construction process is illustrated in Figure~\ref{fig:pipeline}, which we present in more detail below.

\subsection{Base Dataset}
The goal of our evaluation is to assess the application of quality metrics in law as a high stakes domain using a realistic set of tasks. To that end, we collect tasks from Harvey's BigLaw Bench. BigLaw Bench is a privately held (non-public) dataset compiled by 
Harvey \citep{harvey_team_introducing_2024} in 2024. Its tasks were developed by formerly practicing BigLaw attorneys to reflect hypothetical, representative work products of the kind performed in litigation and transactional practice. The documents associated with the released tasks are public, and the tasks are designed to exemplify professional legal work. BigLaw Bench's ability to capture and model realistic processes of high economic value makes it a particularly suitable task collection for our purposes. Each BigLaw Bench task consists of a \textit{(prompt--documents--rubric)} triplet.

\begin{itemize}
    \item The \textit{prompt} describes the legal task and can be used to instruct an LLM to generate a corresponding legal work product.
    \item The \textit{documents} consist of attachments that may be relevant to answering the prompt.
    \item The \textit{rubric} consists of atomistic criteria that can be used to compute a total point value for a given model response. Rubrics contain affirmative criteria, awarding points for elements of structure, style, and substance, as well as negative criteria, subtracting points for hallucinations, irrelevant material, or other shortcomings. Rubrics are task-specific and were developed by legal experts, then later reviewed by other legal experts.
\end{itemize}

BigLaw Bench consists of 100 triplets. All triplets have a prompt and a rubric, but some do not have a document as the associated legal task may not require one. For the purposes of the present study, we sampled 30 tasks from the proprietary, full dataset and make them publicly available.\footnote{We curated our sample to balance perceived complexity, practice area, and task type in order to ensure coverage across the whole spectrum of legal work products.}

\begin{table}[!t]
  \caption{Sample tasks by category and task type.}
  \label{tab:sampled-tasks-by-type}
  \scriptsize
  \centering
  \setlength{\tabcolsep}{3.5pt}
  \renewcommand{\arraystretch}{1.08}
  \begin{tabularx}{\textwidth}{@{}
    >{\raggedright\arraybackslash}p{0.14\textwidth}
    >{\raggedright\arraybackslash}p{0.23\textwidth}
    >{\raggedright\arraybackslash}X
    >{\centering\arraybackslash}p{0.05\textwidth}
    @{}}
    \toprule
    Category & Task Type & Task Example & Count \\
    \midrule
    Litigation & Analysis of Litigation Filings
      & Analyze the Court's order in the motion to dismiss including analysis of remaining claims and potential future arguments.
      & 3 \\
    Litigation & Case Law Research
      & Analyze treatment of motions in limine by different courts and outline any specific filing deadlines.
      & 1 \\
    Litigation & Case Management
      & Analyze complaint and draft questions for initial client interview with the defendant party.
      & 2 \\
    Litigation & Document Review \& Analysis
      & Analyze trial documents and draft an analysis of conflicts, gaps, contradictions, or ambiguities, including a detailed chronology of events and analysis results.
      & 1 \\
    Litigation & Drafting (L)
      & Analyze key arguments in the summary judgment brief and draft counter arguments in response to the arguments made in same.
      & 2 \\
    Litigation & Regulatory \& Advising
      & Draft summary of the Anti-Kickback Statute/s definition of "remuneration" and how it might apply to a client's patient assistance program.
      & 2 \\
    Litigation & Transcript Analysis
      & Analyze deposition transcript and draft a detailed chart of deponent's statements that list and describe the harms alleged verbatim.
      & 1 \\
    Litigation & Trial Preparations \& Oral Argument
      & Analyze trial brief and draft list of question topics and a corresponding list of questions for each topic for the court in its voir dire of jurors.
      & 2 \\
    \addlinespace
    \multicolumn{3}{r}{\textit{Litigation total}}
      & \textit{14} \\
    \midrule
    Transactional & Corporate Strategy \& Advising
      & Draft email correspondence to client summarizing the interim operating covenants and providing guidance on interim operating covenants that restrict the client between signing and closing.
      & 4 \\
    Transactional & Deal Management
      & Draft transaction checklist for an underwritten offering formatted as a chart, detailing relevant parties, action items, and estimated timeline.
      & 2 \\
    Transactional & Drafting (T)
      & Revise indemnification clause in warrant agreement to be more client favorable.
      & 2 \\
    Transactional & Due Diligence
      & Analyze main services agreement for provisions that would be triggered by a change of control of the company.
      & 1 \\
    Transactional & Legal Research
      & Conduct research regarding Delaware corporate law's ratification process to provide an explanation on ratifying company's initial incorporation documents that were not properly approved.
      & 3 \\
    Transactional & Negotiation Strategy
      & Analyze warrant agreements regarding its expiration provisions and provide explanations on expiration mechanics and timing.
      & 1 \\
    Transactional & Risk Assessment \& Compliance
      & Analyze most-favored nation provisions found in side letter and draft a table advising clients on risk of triggering MFN obligations.
      & 1 \\
    Transactional & Transaction Structuring
      & Draft a comparison chart of the three different financing options for a presentation to the board of directors, including immediate action items for each.
      & 2 \\
    \addlinespace
    \multicolumn{3}{r}{\textit{Transactional total}}
      & \textit{16} \\
    \midrule
    \multicolumn{3}{r}{\textbf{Total}}
      & \textbf{30} \\
    \bottomrule
  \end{tabularx}
\end{table}

Table~\ref{tab:sampled-tasks-by-type} summarizes the distribution of tasks in this sample of 30 by category and task type. Appendix Table~\ref{tab:task-level-summary} provides per-task metadata and study metrics. In total, the sample comprises 16 transactional tasks and 14 litigation tasks.

\subsection{Constructing Quality Levels}

We implement quality differences in our output through \textit{construction}. This approach avoids the circularity that comes with defining quality through either of the evaluation methodologies being compared. We use model-generated variation because the intended objects of evaluation are LLM outputs. We contrasted several approaches to induce quality differences, including through various model sizes, thinking budgets and prompt strategies. A practicing attorney with over 30 years of experience in transactional law manually reviewed the different approaches and determined that prompt-induced quality differences are a realistic reflection of the type of variation that might be seen in associate work product. Additionally, we used the expert's feedback over two iterations to refine the response formatting, aligning with senior lawyers' expectations. Ultimately, for each task, we generated work products at three quality levels using Anthropic's Claude Opus 4.6. Any reference documents were chunked and embedded using the Qwen-3 embedding model \citep{zhang2025qwen3embeddingadvancingtext}; for each task, the top 10 chunks most similar to the task prompt were retrieved and included in the model's context alongside the prompt.
To induce variation in output quality, we used a system prompt specifying a profile along six dimensions:
\begin{itemize}
    \item \textbf{Analytical depth:} Analysis of implications and second-order effects.
    \item \textbf{Precision:} Engagement with exact wording rather than paraphrase.
    \item \textbf{Completeness:} Coverage of relevant points.
    \item \textbf{Reasoning clarity:} Explicitness of the logic chain.
    \item \textbf{Judgment:} Willingness to take positions.
    \item \textbf{Nuance:} Handling of ambiguity and competing interpretations.
\end{itemize}
We varied quality by setting all six dimensions to the same degree: \textit{excellent}, \textit{good}, or \textit{intermediate}. The prompt template is included in Appendix~\ref{app:prompt-generation}.
In addition to the qualitative assessment of our expert, we used an LLM-as-a-Judge to confirm that the outputs are qualitatively distinguishable. When prompted (Appendix~\ref{app:prompt-quality-validation}) to identify the better output, OpenAI's GPT-5.4 preferred the excellent output to the good output 69 of 100 times in a within-task pairwise setup, and preferred the good output to the intermediate output 67 of 100 times. Some alternative protocols (e.g. different model sizes) yielded agreement rates of up to 90\%, but we chose a more ambiguous protocol to reduce the risk of making comparative judgments trivial and inducing a ceiling effect. But despite the combined evidence from the expert and the LLM validation, we acknowledge that prompt-induced variation could differ from naturally occurring variation in human work product in important respects. We discuss these limitations and their implications for our findings in Appendix~\ref{app:limitations}.

To reduce sensitivity to sampling variation from any single LLM generation, we generated $N=50$ independent output variants for each task--quality-level cell. That is, for each task and each constructed quality level, we produced 50 separate LLM outputs using the corresponding quality-controlled prompt. We randomize these at the annotator-level, so that, for a given task, two annotators may receive different representations of an \textit{excellent} output, although outputs are consistent within annotator.

\FloatBarrier
\subsection{Annotation Dataset}
We recruited 51 expert lawyers from two sources to annotate our data. Our first set of 11 experts consists of attorneys currently working at one of two large law firms headquartered in the U.S. These attorneys contributed their quality assessments through their firms' \textit{pro bono} programs. The remaining 40 were recruited through a professional AI data lab, Snorkel AI. These practicing attorneys generally have more heterogeneous backgrounds, although all of them have substantive experience in legal practice. Across all the experts, reported years of experience range from 3 to 35 years, with a median of 10. Table~\ref{tab:annotator-information} shows summary information for the contributing annotators. A combined 62.7\% of annotators reported eight or more years of practice (29.4\% with 8–11 years; 13.7\% with 12–15 years; 5.9\% with 16–19 years; 13.7\% with 20 or more years). 37.3\% of annotators were senior associates, 11.8\% were partners, and 9.8\% held counsel positions, indicating that a substantive majority qualified as senior attorneys under prevailing industry conventions. Their practice expertise was largely concentrated in four domains: Litigation (70.6\%), Transactions (62.7\%), Regulatory (43.1\%), and Labor \& Employment (33.3\%). Annotators were permitted to select multiple practice areas; percentages therefore do not sum to 100\%.

Every participating lawyer was assigned a pipeline of all 30 tasks in a random order. For each task in a pipeline, we first sampled one output of $N=50$ output variants for each of the three constructed quality levels. With these three outputs, we then defined two evaluation protocols, rubrics and comparative judgments. Each task required three assessments, presented in random order. For rubrics, annotators scored the sampled outputs for each quality level (\textit{excellent}, \textit{good}, \textit{intermediate}) according to the expert-developed rubric. For comparative judgments, annotators documented their pairwise preference for the three unique pairs formed by those same sampled outputs. The pipeline would randomly initialize with either the rubric or comparative judgment methodology. Every successive task would then alternate between the two methodologies, such that annotators would do a task either under the rubric protocol or the comparative judgment protocol, but not both. Annotators were allowed to skip tasks outside of their area of expertise, but were not able to skip a methodology. For instance, if the first task involved comparative judgments, the next task would need to be completed via rubric, irrespective of whether and how many tasks the annotator chose to skip after the first. The annotators used a custom-designed web interface to advance through tasks. They were encouraged to spend no more than 20 minutes on an individual evaluation, or one hour on the three evaluations comprising a task. Our full annotator instructions are included in Appendix ~\ref{app:human-subjects}.

Overall, the human annotation dataset contains 1,539 rubric scores and 1,530 pairwise preference judgments over 2,274 unique outputs for 30 tasks. Annotators spent a total of 242 hours and 15 minutes on completed analyzed evaluations after excluding 56 records with recorded times over 30 minutes from the time calculation only. Valuing partner-firm time at \$1,000 per hour and Snorkel AI annotator time at \$125 per hour yields an estimated equivalent value of \$48,500. This estimate includes \textit{pro bono} contributions and does not represent actual expenditure.

\section{Results}
\label{sec:results}
We are interested in comparing the extent to which rubric-based scoring and comparative judgments can recover the constructed quality differences in our outputs. We report two estimands, each defined as the cross-method difference under a different recovery metric. The first, $D_{\rho} = \bar{R}^{\mathrm{CJ}} - \bar{R}^{\mathrm{rub}}$, uses the task-averaged Spearman's rank correlation, computed after pooling annotators' judgments within each task between (i) constructed rankings and (ii) rankings implied by the respective method (Bradley-Terry latent utilities for comparative judgments; average rubric scores for rubrics). The second, $D_{W} = \bar{W}^{\mathrm{CJ}} - \bar{W}^{\mathrm{rub}}$, uses the task-averaged per-judgment pairwise win rate: the share of pairwise comparisons across quality levels whose preferred or higher-scored output is the one at the higher quality level (ties, which are only possible in the rubric protocol, count as 0.5). The two estimands differ in their unit of aggregation: $D_{\rho}$ pools individual judgments within each task-method combination before computing rank correlation, while $D_{W}$ averages per-judgment correctness directly. Positive values of either estimand indicate better recovery under comparative judgments. To assess statistical significance, we perform $B = 2{,}000$ hierarchical bootstrap iterations and report 95\% normal-theory confidence intervals centered on the observed estimates; both estimands are recomputed on the same resampled draws each replicate. Details about our methodology appear in Appendix~\ref{app:detailed-methodology}, and the bootstrap distributions appear in Appendix Figure~\ref{fig:bootstrap-d-hat}. A discussion of the substantive interpretation of each estimand can be found in Appendix~\ref{app:interpretation-estimands}.
\subsection{Human Annotators}
\begin{figure}[ht]
  \centering
  \includegraphics[width=0.85\textwidth]{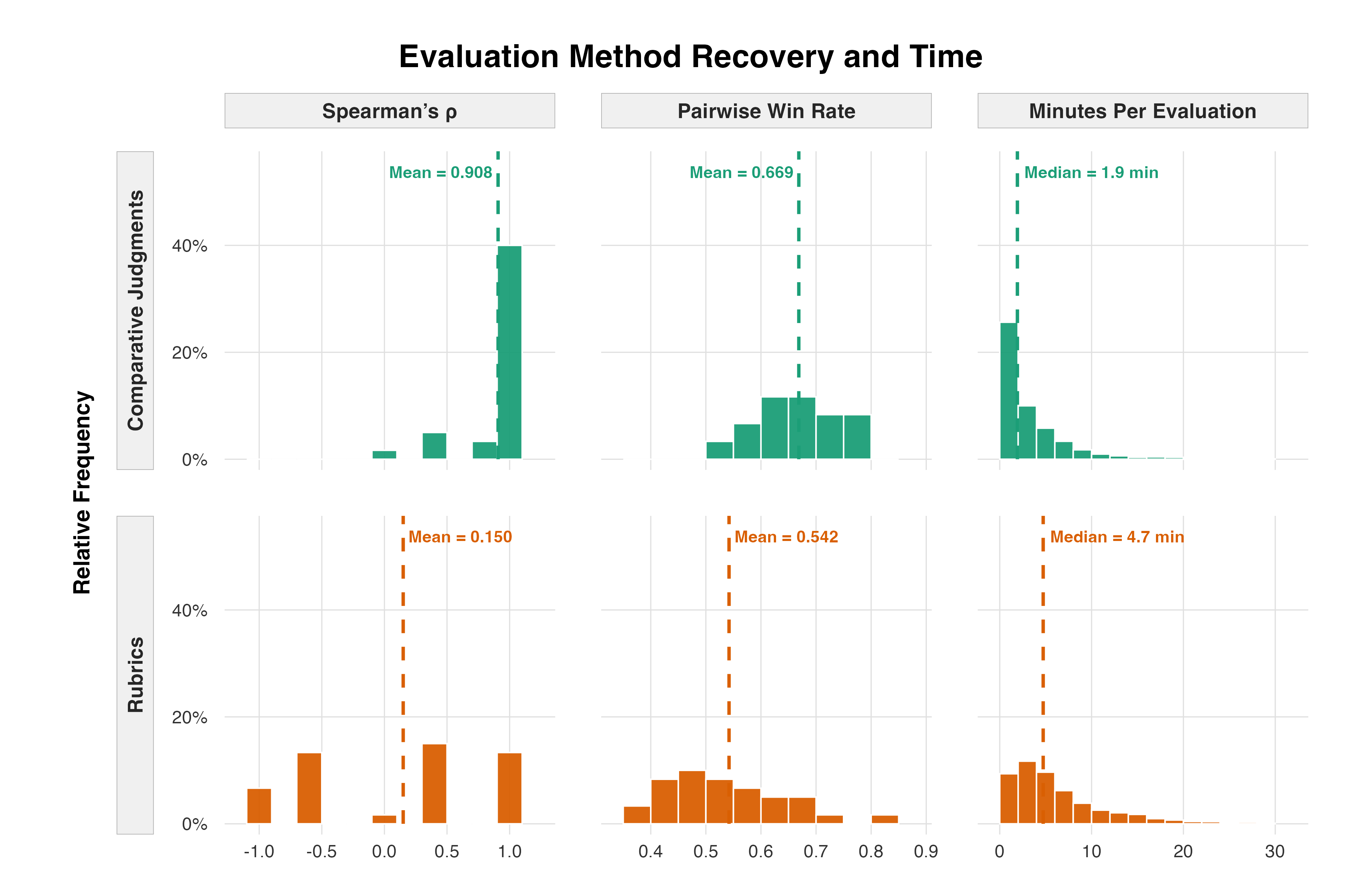}
  \caption{Main recovery and timing results. The first two columns show recovery under two metrics: task-level Spearman's $\rho$ between the constructed quality-level ordering and the method-implied ordering, and the task-level pairwise win rate. To enhance legibility, timing results in the third column omit 56 outliers with recorded times between 30.1 and 3,244.1 minutes.}
  \label{fig:main-results}
\end{figure}

We find strong evidence that annotators better surface the constructed quality ordering when they apply comparative judgments over rubrics (Figure~\ref{fig:main-results}; rank-correlation metric $\bar{R}^{\mathrm{CJ}}=0.908$, $\bar{R}^{\mathrm{rub}}=0.150$, $\widehat{D}_{\rho}=0.758 [0.494, 1.021]$; win-rate metric $\bar{W}^{\mathrm{CJ}}=0.669$, $\bar{W}^{\mathrm{rub}}=0.542$, $\widehat{D}_{W}=0.127 [0.067, 0.186]$). The advantage materializes despite comparative judgments requiring much less time than rubrics (median 4.73 minutes for rubrics, 1.92 minutes for comparative judgments). The performance difference persists across varying levels of annotator experience, consistent with prior evidence that novice and experienced raters can produce comparable reliability and validity under relative judgment protocols \citep{sims_rubric_2020}. At the same time, within-annotator estimates exhibit substantial heterogeneity (Appendix Figure~\ref{fig:experience-diagnostic}). The advantage is substantively similar across other subsets of the dataset as well, including task category, annotator experience, and annotation time (Appendix Tables~\ref{tab:subgroup-recovery-summary} and~\ref{tab:subgroup-recovery-summary-winrate}). The advantage also holds for pairs of adjacent quality levels only (Appendix Table~\ref{tab:adjacent-pair-accuracy}). In Figure~\ref{fig:quality-level-positive-point-score-distributions}, we further examine whether the relatively weak performance of rubrics could be explained by the inability of the expert rubrics to measure finer nuances of quality. This would be the case if, for instance, outputs of all quality levels received generally high scores. However, Figure~\ref{fig:quality-level-positive-point-score-distributions} does not suggest a general ceiling effect for human annotators. Instead, positive rubric scores are widely distributed over outputs across the full range of achievable scores. Overall, our results indicate that comparative judgments have a consistent, material advantage over the supplied expert-developed rubrics in recovering our quality manipulation, both in terms of performance and time.

While total rubric scores can vary widely, this does not necessarily mean that the rubrics distinguish between quality levels. We investigated positive points awarded on the level of individual criteria. For each task and criterion, we compared the scores for outputs from different quality levels. For instance, an excellent and an intermediate output could both receive zero points or both receive one point on a criterion. These count as ties, even if the outputs' total rubric scores differ. Approximately three-quarters of these criterion-level comparisons were ties. Among the other comparisons, the higher-quality output more often received more points.

We also checked whether penalties explain the weak recovery of the constructed ordering. We removed negative deductions from the total scores and recomputed the rubric rankings. This made little difference to human rubric recovery (Appendix~\ref{app:additional-analyses}).

We also examined whether comparative judgments were consistent across annotators. For each judgment, we compared the annotator's preferred quality level with the majority choice of the other annotators evaluating the same task and quality comparison, such as excellent versus good. They agreed in 66.2\% of cases (annotator-clustered bootstrap 95\% CI: 61.8--70.7\%). Annotators generally saw different versions of the outputs, so this measures agreement about quality levels rather than identical answers.

We separately checked whether any one annotator drove the overall preference for higher-quality outputs. Removing each annotator's judgments in turn changed the higher-quality win rate from 66.9\% to between 66.1\% and 67.5\% (Appendix~\ref{app:additional-analyses}).

In our dataset, constructed quality is strongly correlated with length (Appendix~\ref{app:length-inversion}). This correlation could be inherent. After all, quality in law often is expressed in the consideration of edge cases, nuanced reasoning and attention to detail, all of which might necessarily result in longer responses. At the same time, it has been documented that human preference ranks can be subject to length biases \citep{chen_humans_2024,fleckenstein2020length}. To assess the potential impact of length on preference ranks, we supplement our analysis with 300 matched comparisons in which returning annotators judged pairs after we reversed which output was longer by regenerating the lower-quality output. The symmetric model assumes that being longer helps the excellent output as much as being shorter hurts it, while the effects of other differences between the outputs remain unchanged. Under this assumption, approximately 66\% of the original preference advantage above chance is associated with length. We again emphasize that these estimates cannot necessarily be understood as length biases, given a potential, inherent correlation between length and quality.

\subsection{Autograders}
Increasingly, investigators utilize LLM autograders to perform evaluations \citep{zheng_judging_2023}. To complement the human annotation dataset, we further employed GPT-5.4 to perform rubric-based scoring and comparative judgments on the same set of completed task-method combinations derived from the human annotation process. We find broadly consistent results. In particular, rankings following comparative judgments better recover our constructed quality rankings under both estimands, overall and within subsets of task categories (Appendix Tables~\ref{tab:subgroup-recovery-summary} and~\ref{tab:subgroup-recovery-summary-winrate}). The mean task-level Spearman's rank correlation between human and GPT-5.4 autograder-implied quality orderings is 0.200 for rubrics and 0.375 for comparative judgments. Results are similarly consistent for the smaller GPT-5.4-mini, with mean task-level Spearman's rank correlations between human and autograder-implied quality orderings of 0.242 for rubrics and 0.611 for comparative judgments. We further expanded the analysis to eight autograders spanning multiple providers and model sizes. Comparative judgments achieve higher per-judgment win rates than rubric scoring for every model tested (Appendix Table~\ref{tab:expanded-autograders}). For GPT-5.4, negative rubric deductions offset the increasing positive-point scores of higher-quality outputs. Removing these deductions substantially improves rubric recovery (Appendix~\ref{app:additional-analyses}). The relevant prompts for the autograders are included in Appendix~\ref{app:prompt-llm-judge-analysis}.

\section{Discussion}
\label{sec:discussion}

Our central finding is that comparative judgments recover constructed quality differences substantially better than the supplied rubrics, and do so in less than half the annotation time. However, the length-inversion results suggest that recovery of the constructed ordering partly reflects the association between quality and length. The lower time cost may also partly reflect reliance on heuristics, such as favoring longer outputs. Two complementary mechanisms are also consistent with this pattern. The first suggests that human raters discriminate more reliably between two stimuli than when assigning absolute values to one item in isolation; relative judgment imposes lower cognitive load and produces more stable signals across raters. The second, more specific to our domain, suggests that legal work product may have tacit, holistic dimensions--analytical depth, persuasive force, strategic judgment--that experienced attorneys recognize but find difficult to enumerate in advance. Rubrics, by constraining attention to predetermined criteria, may systematically miss precisely those dimensions. Our design cannot adjudicate fully between these mechanisms, but two patterns in the data are suggestive. The persistence of the advantage for comparative judgments across annotator experience levels is consistent with the rationale of easing the cognitive load, since less experienced raters benefit even though they have less tacit knowledge to recover. At the same time, the gap may reflect limitations of the supplied rubrics or their alignment with the quality manipulation. Our results do not distinguish these explanations from a broader advantage in capturing holistic quality. We discuss these limitations together in Appendix~\ref{app:limitations}.

We note that our results should not be read as an uncritical endorsement of comparative judgments over rubric-based scoring. The two methodologies produce structurally different signals, and the choice between them may depend on what the signal is for. Annotations derived from comparative judgments collapse into a single ordinal preference, whereas rubrics can offer insights into why one output might be preferred over another. Rubrics also support per-criterion auditing in ways that aggregated preferences cannot, which matters in deployment contexts where stakeholders need to understand what is being measured. This asymmetry suggests a more nuanced reading of our results: the supplied rubrics recover our constructed quality ordering less effectively, but may retain value as an analytical instrument. A natural direction for ongoing and future work, which we flag below, is how and whether the two can be combined.  For instance, rubrics can act as scaffolding that informs comparative judgments rather than as a scoring instrument in their own right. Similarly, recent work has examined whether rubrics can be derived from comparative judgments retroactively without the loss of quality signal \citep{rezaei2025onlinerubricselicitationpairwise,lv_learning_2026,liu_openrubrics_2025,liu2026cdrrmcontrastdrivenrubricgeneration}. Both directions appear to be particularly fruitful grounds for further progress.

Beyond evaluation, our findings speak to an open question in the post-training literature: whether rubric-based or preference-based supervision is better suited to domains without verifiable rewards. Recent work has argued for rubrics and rubric-style decomposed reward signals on the grounds that they produce more interpretable and auditable reward signals \citep{dineen-etal-2025-qa} and, in some settings, more sample-efficient reward signals than preference-trained reward models \citep{viswanathan2025checklistsbetterrewardmodels}, with rubric-trained policies outperforming both direct Likert and reference-based judging in medicine and science \citep{gunjal2025rubricsrewardsreinforcementlearning}. A maintained assumption of much of this work is that rubrics generated from reference answers are roughly as informative as rubrics authored by domain experts--a finding established within domains in which reference answers are unusually clean. Whether the same conclusion holds in judgment-rich professional domains is an open empirical question that paired rubric–preference data is structured to answer. We focus on evaluation in this paper; we do not attempt the training comparison, in order to keep the present contribution focused on the methodological question and the resource that supports it.

In the future, researchers can apply alternative rubrics to the released outputs and compare their recovery of expert preferences with that of the supplied rubrics. They can also evaluate LLM judges against either expert signal, measuring agreement with pairwise preferences or correspondence with criterion-level rubric labels. For post-training research, separate reward models can be fitted using each signal and compared on held-out tasks. Separating tasks across development and evaluation avoids leakage between output variants.

Two further observations are worth surfacing. First, the time cost of comparative judgment is roughly 40\% of that of the rubric protocol in our data, materially changing the cost-benefit calculus for expert-annotated benchmarks. Together with the greater recovery of our constructed ordering, this finding makes comparative judgment a promising option for similar legal evaluations, while rubrics may remain useful where criterion-level diagnostic information is required. Second, the advantage of comparative judgments is substantively present across annotator experience groups, although within-annotator estimates exhibit substantial heterogeneity, and the observed win-rate advantage is least pronounced among the most senior practitioners in our study.

\section*{Acknowledgments}
This project was supported by our data partner, Snorkel AI.

This project was also supported by Harvey, who provided the tasks and associated data.

We thank our law firm partners, including Vorys, Sater, Seymour and Pease LLP, for contributing lawyer annotations through their pro bono programs. We are especially grateful to Scott Powell at Vorys for his support.

\bibliographystyle{plainnat}
\bibliography{references}

\clearpage
\appendix

\section{Technical appendices and supplementary material}

\subsection{Limitations}
\label{app:limitations}
The present study should be interpreted in light of several limitations, primarily to its external validity and our interpretation of the findings.

Constructed quality is strongly associated with output length, which complicates the interpretation of comparative judgments. A follow-up analysis with 300 matched judgments found that regenerating the lower-quality output to make it longer than the unchanged excellent output substantially reduced preference for excellent outputs. Under a symmetric model, approximately 66\% of the original advantage above chance is associated with length. Given that length might be correlated with nuance and attention to detail, this association may reflect substantive quality differences, but it could also reflect length bias. Appendix~\ref{app:length-inversion} reports the design, results, and assumptions.

Our design also relies on inducing of quality variation through prompting. Prompt-induced variation has two features that may not generalize. First, by construction, our quality manipulation moves all six dimensions of the quality profile together; real variation in attorney work product is more likely to be mixed, with a brief that is strong on analytical depth but weak on precision, or thorough but unclear. An additional experiment with mixed quality profiles yields a similar method difference for autograders (Appendix~\ref{app:additional-analyses}), but does not establish generalization to naturally occurring trade-offs or human evaluation of such profiles. Second, prompt-induced variation may produce stylistic markers that correlate with quality level and that holistic judgment can detect even without engaging with substance. Although our domain expert qualitatively validated the realism of the constructed outputs, we cannot rule out either possibility. One natural alternative would have been to construct quality variation from human-authored work product at different experience levels. However, such an approach faces practical and conceptual difficulties. Practically, obtaining calibrated human-authored outputs across multiple quality levels for 30 real-world legal tasks would require substantial additional expert labor, and quality calibration across human authors itself requires a way to measure quality--raising the exact methodological concerns we seek to address. Conceptually, benchmarks of this kind are increasingly used to evaluate model rather than human outputs, and the variation that matters for those evaluations is variation produced by models. Human-authored variation may, in that sense, be a less appropriate target than the prompt-induced variation we use, even if it could be argued to be more ``natural.'' We nonetheless acknowledge that the magnitude of the advantage in favor of comparative judgments we observe may shrink in settings where quality variation is multidimensional or where surface cues are less informative than they are in our outputs.

Furthermore, although the rubrics we used were generated by experts, it is possible that a different set of rubrics would increase the relative strength of the corresponding methodology. External validity is also limited through task construction. The tasks we included are all derived from BigLaw Bench. As such, although they arguably provide a valuable and valid representation of work processes in transactional and litigation groups at large law firms, we do not include tasks reflective of other important areas of practice, including criminal law, constitutional law, antitrust law, family law, immigration, regulatory affairs, or other specialized practices. The tasks also do not cover dynamic legal tasks, such as taking a deposition or negotiating a live deal. Relatedly, the base dataset was constructed with a distribution of tasks mirroring time spent on billable work, but some modes of important work in a lawyer's professional life--such as client development, mentorship, or supervisory matters--are difficult to enumerate in a timesheet, further limiting the generalizability of these findings.

In addition to these substantive limitations on external validity, we note that our study is geographically restricted to the U.S. Both the underlying tasks and the annotators reflect the conventions of U.S. common-law practice. Civil-law jurisdictions, jurisdictions with different drafting and pleading conventions, and legal systems organized around different professional structures may yield different results, and our findings should not be assumed to generalize to non-U.S. legal contexts without further validation.

We also note that, in generating our dataset, we opted to generate 50 outputs per task per quality level. In effect, this means that annotators rarely saw and scored the exact same output, preventing us from examining the IRR holistically. Overlapping items were assessed by multiple annotators only 206 times for rubrics and 3 times for comparative judgments, due to their pairwise structure. Within rubrics, the IRR was 0.442, suggesting moderate agreement. For comparative judgments, we report stability across annotators within task--quality-pair strata (Appendix~\ref{app:additional-analyses}).

\subsection{Length and Comparative Judgments}
\label{app:length-inversion}

\paragraph{Length in the original data.}
Constructed quality is closely associated with length in the generated outputs. In the original human comparisons, picking the longer output is equivalent to picking the higher quality output in 99.3\% of cases, versus a 66.9\% win rate for the human preference judgments.

Mean output length was highest in excellent outputs, followed by good outputs, and then intermediate outputs in all 30 tasks. The median task-level ratios are 1.38 for excellent to good, 1.41 for good to intermediate, and 1.99 for excellent to intermediate.

Table~\ref{tab:length-gap-quartiles} reports human recovery rates by the absolute word-count gap between the two outputs.

\begin{table}[ht]
\centering
\small
\caption{Human preference win rate by absolute word-count gap.}
\label{tab:length-gap-quartiles}
\begin{tabular}{lrrr}
\toprule
Quartile & Word-count gap & Comparisons & Recovery rate \\
\midrule
1 (smallest gap) & 3--337 & 383 & 64.0\% \\
2 & 337--595 & 383 & 67.1\% \\
3 & 596--961 & 382 & 70.7\% \\
4 (largest gap) & 962--3,171 & 382 & 65.7\% \\
\bottomrule
\end{tabular}
\end{table}

We did not find evidence that a larger length gap further increased the probability of preferring the output with the higher quality ($p=0.206$, from an unclustered logistic regression on the log of one plus the absolute word-count gap). This differs from the follow-up, which reverses which output is longer. The findings are therefore compatible with judges responding to the direction of the length difference more than its magnitude.

\paragraph{Length inversion.}
To examine the role of length, returning annotators reevaluated pairs they had judged in the original study. For each pair, we:
\begin{itemize}
    \item Kept the original excellent output.
    \item Set the target word count for the new intermediate or good output to $T=2w_E-w_B$, where $w_E$ and $w_B$ are the original word counts.
    \item Generated a new output using the same model and settings, adding to the prompt ``Produce a response in approximately [requested] words, within $\pm2\%$ of that target. Do not mention this length requirement.'' Initially, the requested count was $T$.
    \item Accepted the first complete output within $\pm5\%$ of $T$ that was longer than the excellent output.
    \item Otherwise, multiplied the requested count by $T$ divided by the generated word count, bounded the multiplier to $[0.8,1.25]$, rounded to the nearest integer, and retried.
\end{itemize}

Original annotators returned to the study and were presented with the original excellent output and the new long output in randomized order. Assignments prioritized excellent--intermediate comparisons followed by excellent--good comparisons. Collection stopped at 300 completed judgments from 12 annotators: 180 excellent--intermediate and 120 excellent--good comparisons.

The excellent-output win rate fell from 70.3\% to 43.7\% (Table~\ref{tab:length-inversion-results}).

\begin{table}[ht]
\centering
\small
\caption{Human preference win rates before and after length inversion, matched within annotator and original pair.}
\label{tab:length-inversion-results}
\begin{tabular}{lrrrr}
\toprule
Comparison & $N$ & Before & After & Drop [95\% CI] \\
\midrule
Excellent vs.\ Intermediate & 180 & 74.4\% & 42.8\% & 31.7 [14.6, 46.9] \\
Excellent vs.\ Good & 120 & 64.2\% & 45.0\% & 19.2 [4.3, 33.8] \\
Overall & 300 & 70.3\% & 43.7\% & 26.7 [11.3, 41.1] \\
\bottomrule
\end{tabular}
\end{table}

\paragraph{Symmetric decomposition.}
We considered a simple model where inversion reverses the length contribution only:
\[
p_{\mathrm{before}}=0.5+Q+L,
\qquad
p_{\mathrm{after}}=0.5+Q-L.
\]
Here, $L$ denotes the length-associated component and $Q$ the remaining component.
\[
L=\frac{p_{\mathrm{before}}-p_{\mathrm{after}}}{2},
\qquad
Q=\frac{p_{\mathrm{before}}+p_{\mathrm{after}}}{2}-0.5.
\]

Table~\ref{tab:length-decomposition} reports the decomposition by comparison type, with confidence intervals for the pooled estimates.

\begin{table}[ht]
\centering
\small
\caption{Symmetric-model decomposition by comparison type. Components are in percentage points; original advantage is measured above 50\%.}
\label{tab:length-decomposition}
\begin{tabular}{lrrr}
\toprule
Quantity & Excellent vs.\ Intermediate & Excellent vs.\ Good & Overall \\
\midrule
Original advantage & 24.4 pp & 14.2 pp & 20.3 pp \\
Length-associated component ($L$) & 15.8 pp & 9.6 pp & 13.3 pp \\
Remaining advantage ($Q$) & 8.6 pp & 4.6 pp & 7.0 pp \\
Implied length-balanced win rate & 58.6\% & 54.6\% & 57.0\% \\
Remaining share & 35.2\% & 32.4\% & 34.4\% \\
\midrule
95\% CI for $L$ & & & [5.7, 20.5] pp \\
95\% CI for $Q$ & & & [0.7, 13.2] pp \\
95\% CI for remaining share & & & [5.8\%, 56.5\%] \\
\bottomrule
\end{tabular}
\end{table}

The confidence intervals were computed with 200,000 percentile bootstrap resamples of annotators, preserving their matched before/after judgments. However, they do not account for dependence within tasks.

The decomposition does not identify length bias separately from substantive quality. Longer outputs may contain more useful analysis, while regeneration and selection for the word-count target may alter quality beyond length.

\subsection{Detailed Methodology}
\label{app:detailed-methodology}

Our analysis compares the performance of rubric-based and comparative judgment-based evaluations. Our primary interest lies in the ability of rubric-implied rankings and comparative judgment-implied rankings to recover the constructed quality rankings across tasks.

Let
\[
\mathcal{Q} = \{I,G,E\}
\]
denote the three levels of quality, corresponding to \textit{intermediate}, \textit{good}, and \textit{excellent}. We define the constructed quality order as intermediate $<$ good $<$ excellent, i.e.
\[
\boldsymbol{g} = (g_I,g_G,g_E) = (1,2,3).
\]
For each task and method, we aggregate all completed evaluations to calculate the implied, average strength of outputs at each quality level. For rubrics, the implied strength at a given quality level is simply the average score assigned to the corresponding outputs:
\[
\widehat{\mu}^{\mathrm{rub}}_{t,q}
=
\frac{1}{n^{\mathrm{rub}}_{t,q}}
\sum_{i=1}^{n^{\mathrm{rub}}_{t,q}}
s^{\mathrm{rub}}_{i,t,q},
\]
where $\widehat{\mu}^{\mathrm{rub}}_{t,q}$ is the pooled mean rubric strength at quality level $q \in \mathcal{Q}$ on task $t$, $s^{\mathrm{rub}}_{i,t,q}$ is the rubric score assigned in completed rubric evaluation $i$, and $n^{\mathrm{rub}}_{t,q}$ is the number of completed rubric evaluations for task $t$ and quality level $q$.

For the comparative judgment protocol, each completed evaluation yielded three pairwise preference judgments. Following standard practice \citep{pollitt_method_2012,kinnear_comparative_2025}, we aggregate these pairwise judgments at the task level using a Bradley-Terry model \citep{bradley_rank_1952}:
\[
\Pr(q \succ q' \mid t)
=
\frac{\exp(\theta_{t,q})}
{\exp(\theta_{t,q})+\exp(\theta_{t,q'})},
\qquad
q,q' \in \mathcal{Q}, \; q \neq q',
\]
with the identifying constraint
\[
\sum_{q \in \mathcal{Q}} \theta_{t,q}=0.
\]
For the comparative judgment arm, the scores are the latent utilities at each level of quality:
\[
\widehat{\mu}^{\mathrm{CJ}}_{t,q}
=
\widehat{\theta}_{t,q}.
\]

For each task $t$ and method $m \in \{\mathrm{rub},\mathrm{CJ}\}$, task-level recovery is defined as the Spearman's rank correlation between the constructed ordering $\boldsymbol{g}$ and the strength-implied ordering $\widehat{\boldsymbol{\mu}}^m_t$:
\[
R^m_t
=
\rho_{\mathrm{S}}
\left(
\boldsymbol{g},
\widehat{\boldsymbol{\mu}}^m_t
\right),
\]
where
\[
\widehat{\boldsymbol{\mu}}^m_t
=
\left(
\widehat{\mu}^m_{t,I},
\widehat{\mu}^m_{t,G},
\widehat{\mu}^m_{t,E}
\right).
\]

We aggregate this statistic across tasks:
\[
\bar{R}^{m}
=
\frac{1}{|\mathcal{T}|}
\sum_{t \in \mathcal{T}}
R^m_t,
\]
From these task-averaged correlations, the rank-correlation estimand is the cross-method difference:

\[
D_{\rho}
=
\bar{R}^{\mathrm{CJ}}
-
\bar{R}^{\mathrm{rub}}.
\]
Positive values of $D_{\rho}$ indicate better recovery under comparative judgment-based evaluation, while negative values indicate better recovery under rubric-based evaluation.

We additionally report the per-judgment pairwise win rate. Each individual comparative-judgment annotation contributes one binary observation, scored correct when the preferred output is the one at the higher quality level. Each rubric evaluation contributes three pairwise comparisons of $s^{\mathrm{rub}}$ across the three quality levels, scored correct when the higher-scored output is the one at the higher quality level and 0.5 when the two scores tie. For each task $t$ and method $m$, the per-task win rate is the mean correctness across individual judgments, $W^m_t$. The task-averaged win rate is
\[
\bar{W}^{m} = \frac{1}{|\mathcal{T}|} \sum_{t \in \mathcal{T}} W^m_t,
\]
and the win-rate estimand is the cross-method difference,
\[
D_{W} = \bar{W}^{\mathrm{CJ}} - \bar{W}^{\mathrm{rub}}.
\]
Positive values of $D_{W}$ indicate better recovery under comparative judgment-based evaluation. Unlike $D_{\rho}$, the win-rate approach does not pool individual judgments via a per-task model, so $D_{W}$ is invariant in expectation to the number of annotators contributing per task.

To separate the estimand from the sampling uncertainty induced by finite evaluator participation, we also employ a hierarchical bootstrap strategy to validate our findings. For each bootstrap replicate, we resample annotators with replacement, and within each resampled annotator, we then resample the completed tasks with replacement, preserving the three assessments associated with each completed task. We compute $\widehat{D}_{\rho}$ and $\widehat{D}_{W}$ on the same resampled draws each replicate and set the bootstrap iterations to $B = 2{,}000$.  Because $\widehat{D}_{\rho}=\bar{R}^{\mathrm{CJ}}-\bar{R}^{\mathrm{rub}}$ is bounded above (each $\bar{R}^m$ is ceiled at 1), the bootstrap distribution of $\widehat{D}_{\rho}$ is asymmetric. The win-rate bootstrap distribution is more symmetric. For each estimand $\widehat{D} \in \{\widehat{D}_{\rho}, \widehat{D}_{W}\}$, we use the bootstrap only to estimate uncertainty and report a normal-theory confidence interval centered on the observed estimate: \[\left[\widehat{D} - z_{0.975}\,\widehat{\mathrm{SE}}_{\mathrm{boot}}, \; \widehat{D} + z_{0.975}\,\widehat{\mathrm{SE}}_{\mathrm{boot}}\right],\] where $\widehat{\mathrm{SE}}_{\mathrm{boot}}$ is the standard deviation of $\widehat{D}^{*}$ across the $B$ bootstrap replicates.

\subsection{Interpretation of the Estimands}
\label{app:interpretation-estimands}
The two estimands $D_{\rho}$ and $D_{W}$ respond to different practical questions.

The rank-correlation estimand $D_{\rho}$ measures recovery of a pooled signal--the judgments of many annotators aggregated within each task-method combination--and is useful for benchmarking at scale. The win-rate estimand $D_{W}$ measures recovery of the constructed ordering by individual comparisons and is useful when ordering a small fixed set of items once.

The two metrics behave differently as the number of annotators per task grows. For comparative judgments, the pooled per-task estimate converges to the constructed ordering when (i) annotator preferences favor the output at the higher quality level in expectation on each pair across quality levels, and (ii) annotator judgments are sufficiently independent for the sample preference to identify that population preference. Under these conditions, $\bar{R}^{\mathrm{CJ}}$ approaches 1 as more annotators contribute, even when individual accuracy is only modest. For rubrics, the pooled signal also converges, but to the rubric's expected score profile across quality levels, which may not match the constructed ordering. If the rubric's expected score profile does not match the constructed ordering, additional annotation alone will not resolve that discrepancy.

Two corollaries follow. First, a high $\bar{R}^{\mathrm{CJ}}$ does not necessarily mean that individual annotators recover the constructed ordering well. Under the conditions above, aggregation can recover that ordering even when individual discrimination is modest. The win rate gives the more direct read on per-judgment recovery. Second, either methodology can fail when evaluations systematically depart from the constructed ordering. For rubric scoring, the criteria provide an additional shared source of error if they do not capture the manipulated dimensions.

\clearpage

\begin{figure}[p]
  \centering
  \includegraphics[width=\textwidth]{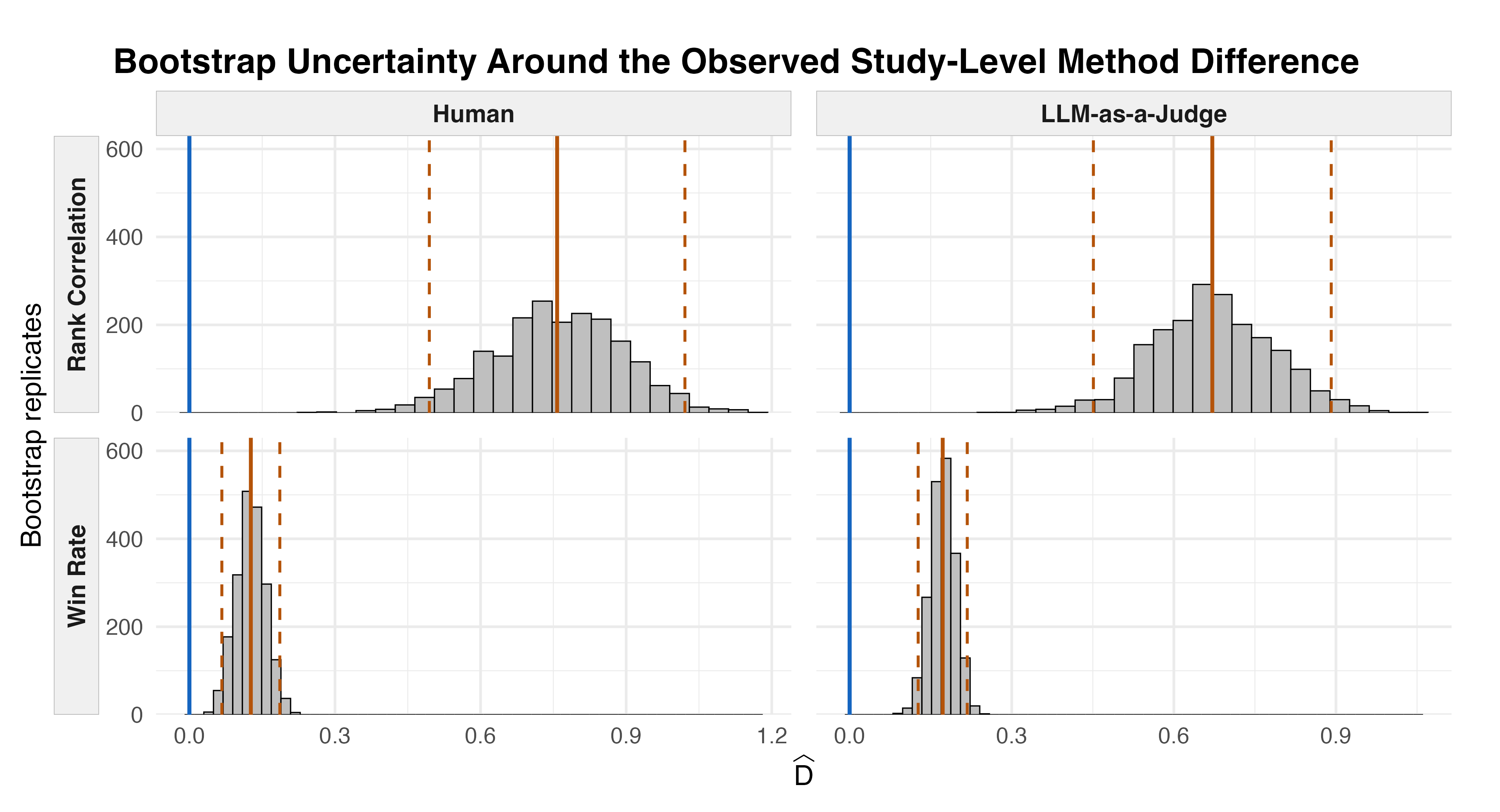}
  \caption{Bootstrap distribution of the study-level method difference. The histograms show $B=2{,}000$ hierarchical bootstrap replicates, where annotators are resampled with replacement and completed task-method evaluations are resampled within annotator. The statistics are $\widehat{D}_{\rho}=\bar{R}^{\mathrm{CJ}}-\bar{R}^{\mathrm{rub}}$ under the rank-correlation metric and $\widehat{D}_{W}=\bar{W}^{\mathrm{CJ}}-\bar{W}^{\mathrm{rub}}$ under the win-rate metric, so positive values favor comparative judgments; the blue vertical line marks $\widehat{D}=0$, or no method difference; dashed vertical lines represent the 95\% confidence intervals: under the rank-correlation metric, $[0.494, 1.021]$ for human annotators and $[0.451, 0.891]$ for autograders; under the win-rate metric, $[0.067, 0.186]$ for human annotators and $[0.127, 0.218]$ for autograders. Because each $\bar{R}$ is ceiled at 1, resampling impacts $\widehat{D}_{\rho}$ asymmetrically. The win-rate bootstrap is bounded similarly but is more symmetric in practice. For both, we limit use of the bootstrap distribution to the estimation of uncertainty and report a normal-theory confidence interval centered on the observed estimate.}
  \label{fig:bootstrap-d-hat}
\end{figure}

\clearpage

\begingroup
\small
\setlength{\tabcolsep}{2pt}
\setlength{\LTleft}{0pt}
\setlength{\LTright}{0pt}
\renewcommand{\arraystretch}{1.16}
\begin{longtable}{@{}
  >{\raggedright\arraybackslash}p{0.420\textwidth}
  >{\raggedright\arraybackslash}p{0.140\textwidth}
  >{\raggedright\arraybackslash}p{0.180\textwidth}
  >{\centering\arraybackslash}p{0.075\textwidth}
  >{\centering\arraybackslash}p{0.055\textwidth}
  >{\centering\arraybackslash}p{0.065\textwidth}
  @{}}
  \caption{Benchmark tasks. $\widetilde{t}$ is median completion time in minutes; skip rate is the share of reached task-method evaluations that were fully skipped.}
  \label{tab:task-level-summary} \\
    \toprule
    Task & Category & Type & \shortstack{$\widetilde{t}$\\(min)} & \shortstack{Skip\\rate} & \shortstack{Rubric\\criteria} \\
    \midrule
  \endfirsthead
  \caption[]{Benchmark tasks. $\widetilde{t}$ is median completion time in minutes; skip rate is the share of reached task-method evaluations that were fully skipped. (continued)} \\
    \toprule
    Task & Category & Type & \shortstack{$\widetilde{t}$\\(min)} & \shortstack{Skip\\rate} & \shortstack{Rubric\\criteria} \\
    \midrule
  \endhead
    \midrule
    \multicolumn{6}{r}{\emph{Continued on next page}} \\
  \endfoot
    \bottomrule
  \endlastfoot
    Analyze most-favored nation provisions found in side letter and draft a table advising clients on risk of triggering MFN obligations. & Transactional & Risk Assessment \& Compliance & 13.1 & 17.9\% & 7 \\
    Draft email correspondence to client summarizing the interim operating covenants and providing guidance on interim operating covenants that restrict the client between signing and closing. & Transactional & Corporate Strategy \& Advising & 15.0 & 8.3\% & 26 \\
    Analyze trial documents and draft an analysis of conflicts, gaps, contradictions, or ambiguities, including a detailed chronology of events and analysis results. & Litigation & Document Review \& Analysis & 18.4 & 0.0\% & 12 \\
    Draft summary of the Anti-Kickback Statute/s definition of "remuneration" and how it might apply to a client's patient assistance program. & Litigation & Regulatory \& Advising & 11.9 & 2.6\% & 12 \\
    Analyze key arguments in the summary judgment brief and draft counter arguments in response to the arguments made in same. & Litigation & Drafting (L) & 21.6 & 7.5\% & 22 \\
    Analyze the Court's order in the motion to dismiss including analysis of remaining claims and potential future arguments. & Litigation & Analysis of Litigation Filings & 11.6 & 5.9\% & 12 \\
    Draft a memorandum to Ansys, Inc.'s general counsel regarding the treatment of employee equity awards in the merger transaction. & Transactional & Corporate Strategy \& Advising & 15.1 & 18.9\% & 23 \\
    Analyze defendant's response to plaintiff's interrogatories including an analysis on admissions and nonresponsiveness. & Litigation & Analysis of Litigation Filings & 13.3 & 5.0\% & 8 \\
    Analyze complaint and draft questions for initial client interview with the defendant party. & Litigation & Case Management & 11.6 & 2.5\% & 15 \\
    Analyze trial brief and draft list of question topics and a corresponding list of questions for each topic for the court in its voir dire of jurors. & Litigation & Trial Preparations \& Oral Argument & 18.9 & 2.9\% & 15 \\
    Analyze a motion to dismiss and response to motion to dismiss and draft outline of reply brief, including main claims and responses to plaintiff's counter arguments. & Litigation & Drafting (L) & 18.6 & 7.9\% & 22 \\
    Analyze treatment of motions in limine by different courts and outline any specific filing deadlines. & Litigation & Case Law Research & 13.7 & 0.0\% & 12 \\
    Conduct research in the company's articles of incorporation regarding requisite shareholder approvals for modifying rights for different classes of shares. & Transactional & Corporate Strategy \& Advising & 9.5 & 10.8\% & 5 \\
    Analyze main services agreement for provisions that would be triggered by a change of control of the company. & Transactional & Due Diligence & 10.5 & 7.9\% & 3 \\
    Draft a comparison chart of the three different financing options for a presentation to the board of directors, including immediate action items for each. & Transactional & Transaction Structuring & 12.2 & 18.4\% & 20 \\
    Draft transaction checklist for an underwritten offering formatted as a chart, detailing relevant parties, action items, and estimated timeline. & Transactional & Deal Management & 14.4 & 15.0\% & 10 \\
    Analyze offering regarding its terms and the existence of a financial advisor fairness opinion and draft a bullet point summary with analysis results. & Transactional & Transaction Structuring & 11.3 & 10.0\% & 14 \\
    Analyze warrant agreements regarding its expiration provisions and provide explanations on expiration mechanics and timing. & Transactional & Negotiation Strategy & 15.9 & 13.5\% & 14 \\
    Revise indemnification clause in warrant agreement to be more client favorable. & Transactional & Drafting (T) & 12.1 & 10.0\% & 6 \\
    Conduct research regarding Delaware corporate law's ratification process to provide an explanation on ratifying company's initial incorporation documents that were not properly approved. & Transactional & Legal Research & 11.7 & 11.4\% & 6 \\
    Conduct research regarding the proper filing form for a proxy statement in which a company proposal is contested. & Transactional & Legal Research & 5.4 & 12.5\% & 2 \\
    Analyze the potential objections to opposing counsel's requests for production and evaluate which objections would be most convincing to a federal court. & Litigation & Case Management & 16.2 & 8.1\% & 24 \\
    Analyze deposition transcript and draft a detailed chart of deponent's statements that list and describe the harms alleged verbatim. & Litigation & Transcript Analysis & 11.6 & 2.6\% & 9 \\
    Conduct research regarding leasehold title policy requirements, including which policies are necessary or recommended. & Transactional & Legal Research & 11.6 & 8.3\% & 3 \\
    Draft email correspondence regarding sections of the draft of a proxy statement explaining its content and its importance, including a request for CEO feedback on same. & Transactional & Corporate Strategy \& Advising & 6.3 & 8.3\% & 6 \\
    Analyze motion for protective order with respect to the court's previous order, including an analysis on evident weaknesses in same. & Litigation & Analysis of Litigation Filings & 16.1 & 5.3\% & 13 \\
    Draft a summary of the credit agreement for disclosure in Item 1.01 of Form 8-K to be filed with the SEC, including standard disclosures and a detailed summary of the negative covenants. & Transactional & Drafting (T) & 16.5 & 22.2\% & 22 \\
    Draft the regulatory risks associated with changing a manufacturer arrangement after a favorable OIG advisory opinion. & Litigation & Regulatory \& Advising & 15.8 & 13.9\% & 18 \\
    Draft an explanation of potential objections to a third-party subpoena in federal court and create a detailed outline of the best objection. & Litigation & Trial Preparations \& Oral Argument & 11.4 & 2.5\% & 12 \\
    Draft a signing and closing checklist outlining all of the key action items required for a stock purchase transaction based on the agreement. & Transactional & Deal Management & 17.7 & 10.3\% & 20 \\
\end{longtable}
\endgroup

\clearpage

\begin{table}[!t]
  \caption{Annotator summary information}
  \label{tab:annotator-information}
  \centering
  \small
  \setlength{\tabcolsep}{6pt}
  \renewcommand{\arraystretch}{1.12}
  \begin{tabularx}{0.72\textwidth}{@{}
    >{\raggedright\arraybackslash}X
    >{\centering\arraybackslash}p{0.08\textwidth}
    >{\centering\arraybackslash}p{0.08\textwidth}
    @{}}
    \toprule
    Characteristic & n & Share \\
    \midrule
    \multicolumn{3}{c}{\textbf{Overall}} \\
    \midrule
    All annotators & 51 & 100.0\% \\
    \midrule
    \multicolumn{3}{c}{\textbf{Firm/source}} \\
    \midrule
    AmLaw 100 firm & 6 & 11.8\% \\
    AmLaw 200 firm (Vorys) & 5 & 9.8\% \\
    AI Data Lab (Snorkel AI) & 40 & 78.4\% \\
    \midrule
    \multicolumn{3}{c}{\textbf{Title}} \\
    \midrule
    Partner & 6 & 11.8\% \\
    Counsel & 5 & 9.8\% \\
    Senior Associate & 19 & 37.3\% \\
    Junior Associate & 11 & 21.6\% \\
    Attorney & 6 & 11.8\% \\
    Other legal roles & 2 & 3.9\% \\
    Not reported & 2 & 3.9\% \\
    \midrule
    \multicolumn{3}{c}{\textbf{Years of experience}} \\
    \midrule
    <=3 & 3 & 5.9\% \\
    4-7 & 14 & 27.5\% \\
    8-11 & 15 & 29.4\% \\
    12-15 & 7 & 13.7\% \\
    16-19 & 3 & 5.9\% \\
    >=20 & 7 & 13.7\% \\
    Not reported & 2 & 3.9\% \\
    \midrule
    \multicolumn{3}{c}{\textbf{Practice area}} \\
    \midrule
    Litigation & 36 & 70.6\% \\
    Transactions & 32 & 62.7\% \\
    Regulatory & 22 & 43.1\% \\
    Labor \& Employment & 17 & 33.3\% \\
    Intellectual Property & 7 & 13.7\% \\
    Tax & 9 & 17.6\% \\
    Other & 14 & 27.5\% \\
    Not reported & 2 & 3.9\% \\
    \bottomrule
  \end{tabularx}
  \vspace{0.35em}
  \begin{minipage}{0.72\textwidth}
    \footnotesize\emph{Note.} Shares use all annotators as the denominator. Practice areas are not mutually exclusive, so practice-area shares may sum to more than 100\%.
  \end{minipage}
\end{table}

\clearpage

\begin{figure}[p]
  \centering
  \includegraphics[width=0.82\textwidth]{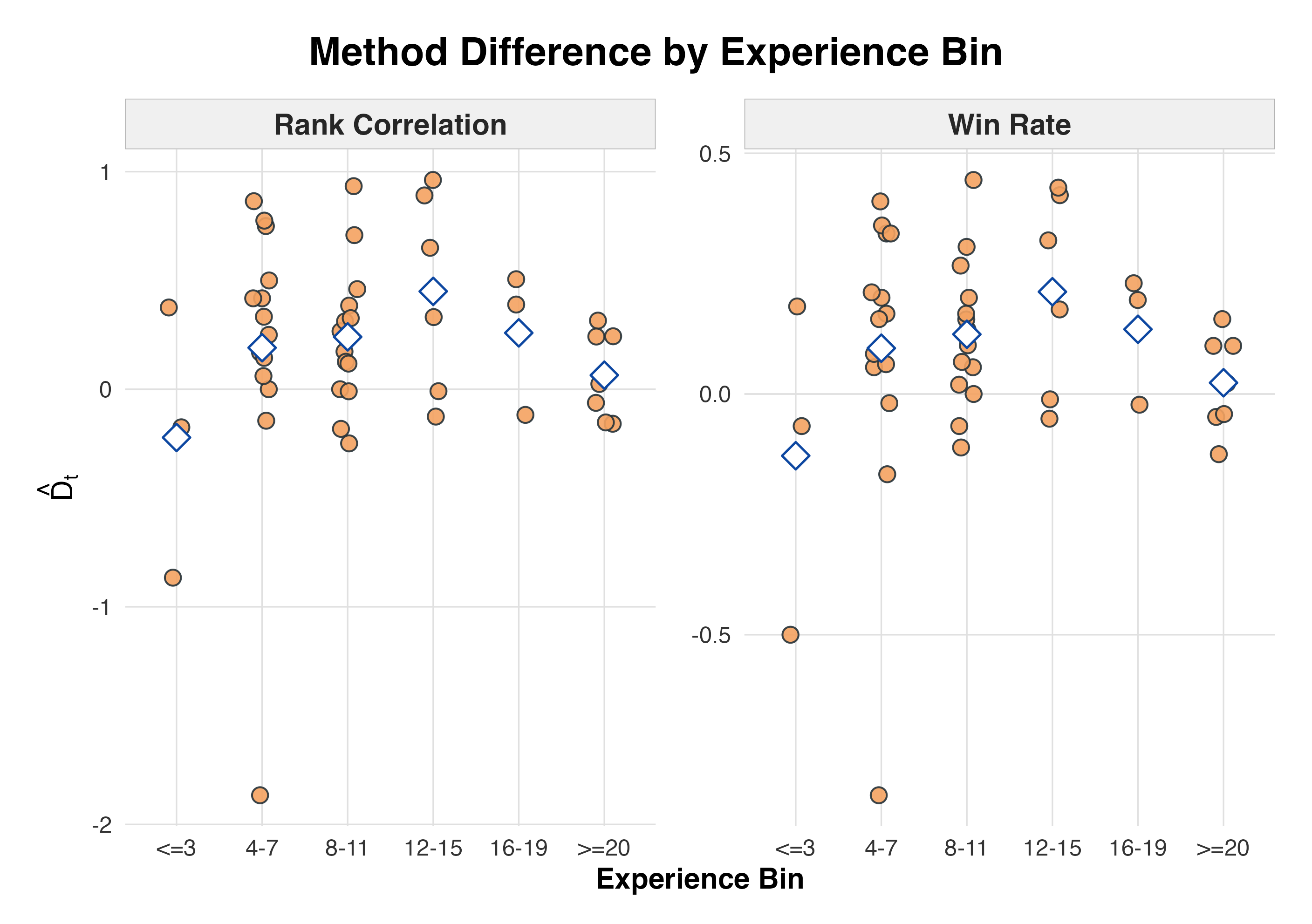}
  \caption{Experience diagnostic. Each point is an annotator with estimable recovery under both methods. The vertical axis is $\widehat{D}_t$, the annotator-level difference between average comparative-judgment performance and average rubric performance; positive values favor comparative judgments.}
  \label{fig:experience-diagnostic}
\end{figure}

\clearpage

\renewcommand{\theHtable}{\thetable}
\renewcommand{\thetable}{7a}
\begin{table}[!t]
  \caption{Subgroup recovery summary. For each subgroup, $\bar{R}^{\mathrm{CJ}}$ and $\bar{R}^{\mathrm{rub}}$ report the mean task-level recovery statistic under comparative judgment and rubric scoring, respectively, where task-level recovery is Spearman's rank correlation between the constructed quality ordering and the method-implied ordering. $\widehat{D}_{\rho}=\bar{R}^{\mathrm{CJ}}-\bar{R}^{\mathrm{rub}}$, so positive values favor comparative judgments. The $n$ columns report complete task-method evaluations contributing to the subgroup before task-level aggregation. GPT-5.4 columns report the corresponding LLM autograder evaluations; the GPT-5.4-mini column reports the method difference for the same full/task subsets. Autograder columns are blank for annotator-defined subgroups.}
  \label{tab:subgroup-recovery-summary}
  \centering
  \scriptsize
  \setlength{\tabcolsep}{1pt}
  \renewcommand{\arraystretch}{1.12}
  \begin{tabular}{@{}
    >{\raggedright\arraybackslash}p{0.101\textwidth}
    >{\raggedright\arraybackslash}p{0.156\textwidth}
    >{\raggedright\arraybackslash}p{0.084\textwidth}
    >{\centering\arraybackslash}p{0.037\textwidth}
    >{\centering\arraybackslash}p{0.057\textwidth}
    >{\centering\arraybackslash}p{0.057\textwidth}
    >{\centering\arraybackslash}p{0.057\textwidth}
    >{\centering\arraybackslash}p{0.057\textwidth}
    >{\centering\arraybackslash}p{0.086\textwidth}
    >{\centering\arraybackslash}p{0.086\textwidth}
    >{\centering\arraybackslash}p{0.070\textwidth}
    >{\centering\arraybackslash}p{0.105\textwidth}
  @{}}
    \toprule
    Subset & Variable & Level & $n$ & \mbox{$\bar{R}^{\mathrm{CJ}}$} & \mbox{$\bar{R}^{\mathrm{rub}}$} & $\widehat{D}_{\rho}$ & \shortstack{GPT-5.4\\$n$} & \shortstack{GPT-5.4\\$\bar{R}^{\mathrm{CJ}}$} & \shortstack{GPT-5.4\\$\bar{R}^{\mathrm{rub}}$} & \shortstack{GPT-5.4\\$\widehat{D}_{\rho}$} & \shortstack{GPT-5.4-mini\\$\widehat{D}_{\rho}$} \\
    \midrule
    Full Dataset & Overall & All & 1023 & 0.908 & 0.150 & 0.758 & 1023 & 0.421 & -0.250 & 0.671 & 0.979 \\
    \addlinespace[0.35em]
    Tasks & Category & Transactions & 524 & 0.858 & -0.094 & 0.952 & 524 & 0.594 & -0.406 & 1.000 & 1.188 \\
    Tasks & Category & Litigation & 499 & 0.964 & 0.429 & 0.536 & 499 & 0.224 & -0.071 & 0.295 & 0.740 \\
    \addlinespace[0.35em]
    Annotators & \mbox{Years of Experience} & $\leq{}3$ & 33 & 0.571 & 0.445 & 0.126 &  &  &  &  &  \\
    Annotators & \mbox{Years of Experience} & 4--7 & 212 & 0.737 & 0.156 & 0.581 &  &  &  &  &  \\
    Annotators & \mbox{Years of Experience} & 8--11 & 361 & 0.728 & 0.267 & 0.461 &  &  &  &  &  \\
    Annotators & \mbox{Years of Experience} & 12--15 & 153 & 0.549 & 0.169 & 0.380 &  &  &  &  &  \\
    Annotators & \mbox{Years of Experience} & 16--19 & 89 & 0.388 & 0.103 & 0.286 &  &  &  &  &  \\
    Annotators & \mbox{Years of Experience} & $\geq{}20$ & 162 & 0.224 & -0.070 & 0.294 &  &  &  &  &  \\
    \addlinespace[0.18em]
    Annotators & Time & $<{}$ median & 511 & 0.703 & 0.001 & 0.702 &  &  &  &  &  \\
    Annotators & Time & $\geq{}$ median & 512 & 0.720 & 0.233 & 0.487 &  &  &  &  &  \\
    \bottomrule
  \end{tabular}
\end{table}

\clearpage

\addtocounter{table}{-1}
\renewcommand{\thetable}{7b}
\begin{table}[!t]
  \caption{Subgroup recovery summary (pairwise win-rate metric). For each subgroup, $\bar{W}^{\mathrm{CJ}}$ and $\bar{W}^{\mathrm{rub}}$ report the mean task-level pairwise win rate under comparative judgment and rubric scoring, respectively, where the per-task win rate is the share of pairwise comparisons across quality levels whose preferred or higher-scored output is the one at the higher quality level (ties, only possible in the rubric protocol, count as 0.5). $\widehat{D}_{W}=\bar{W}^{\mathrm{CJ}}-\bar{W}^{\mathrm{rub}}$, so positive values favor comparative judgments. The $n$ columns report complete task-method evaluations contributing to the subgroup before task-level aggregation. GPT-5.4 columns report the corresponding LLM autograder evaluations; the GPT-5.4-mini column reports the method difference for the same full/task subsets. Autograder columns are blank for annotator-defined subgroups.}
  \label{tab:subgroup-recovery-summary-winrate}
  \centering
  \scriptsize
  \setlength{\tabcolsep}{1pt}
  \renewcommand{\arraystretch}{1.12}
  \begin{tabular}{@{}
    >{\raggedright\arraybackslash}p{0.101\textwidth}
    >{\raggedright\arraybackslash}p{0.156\textwidth}
    >{\raggedright\arraybackslash}p{0.084\textwidth}
    >{\centering\arraybackslash}p{0.037\textwidth}
    >{\centering\arraybackslash}p{0.057\textwidth}
    >{\centering\arraybackslash}p{0.057\textwidth}
    >{\centering\arraybackslash}p{0.057\textwidth}
    >{\centering\arraybackslash}p{0.057\textwidth}
    >{\centering\arraybackslash}p{0.086\textwidth}
    >{\centering\arraybackslash}p{0.086\textwidth}
    >{\centering\arraybackslash}p{0.070\textwidth}
    >{\centering\arraybackslash}p{0.105\textwidth}
  @{}}
    \toprule
    Subset & Variable & Level & $n$ & \mbox{$\bar{W}^{\mathrm{CJ}}$} & \mbox{$\bar{W}^{\mathrm{rub}}$} & $\widehat{D}_{W}$ & \shortstack{GPT-5.4\\$n$} & \shortstack{GPT-5.4\\$\bar{W}^{\mathrm{CJ}}$} & \shortstack{GPT-5.4\\$\bar{W}^{\mathrm{rub}}$} & \shortstack{GPT-5.4\\$\widehat{D}_{W}$} & \shortstack{GPT-5.4-mini\\$\widehat{D}_{W}$} \\
    \midrule
    Full Dataset & Overall & All & 1023 & 0.669 & 0.542 & 0.127 & 1023 & 0.599 & 0.427 & 0.172 & 0.254 \\
    \addlinespace[0.35em]
    Tasks & Category & Transactions & 524 & 0.665 & 0.518 & 0.147 & 524 & 0.616 & 0.357 & 0.259 & 0.296 \\
    Tasks & Category & Litigation & 499 & 0.673 & 0.570 & 0.103 & 499 & 0.581 & 0.508 & 0.073 & 0.205 \\
    \addlinespace[0.35em]
    Annotators & \mbox{Years of Experience} & $\leq{}3$ & 33 & 0.746 & 0.649 & 0.097 &  &  &  &  &  \\
    Annotators & \mbox{Years of Experience} & 4--7 & 212 & 0.743 & 0.555 & 0.187 &  &  &  &  &  \\
    Annotators & \mbox{Years of Experience} & 8--11 & 361 & 0.675 & 0.555 & 0.120 &  &  &  &  &  \\
    Annotators & \mbox{Years of Experience} & 12--15 & 153 & 0.686 & 0.514 & 0.172 &  &  &  &  &  \\
    Annotators & \mbox{Years of Experience} & 16--19 & 89 & 0.637 & 0.550 & 0.086 &  &  &  &  &  \\
    Annotators & \mbox{Years of Experience} & $\geq{}20$ & 162 & 0.557 & 0.506 & 0.050 &  &  &  &  &  \\
    \addlinespace[0.18em]
    Annotators & Time & $<{}$ median & 511 & 0.649 & 0.537 & 0.113 &  &  &  &  &  \\
    Annotators & Time & $\geq{}$ median & 512 & 0.708 & 0.544 & 0.164 &  &  &  &  &  \\
    \bottomrule
  \end{tabular}
\end{table}

\clearpage

\begin{figure}[p]
  \centering
  \includegraphics[width=\textwidth]{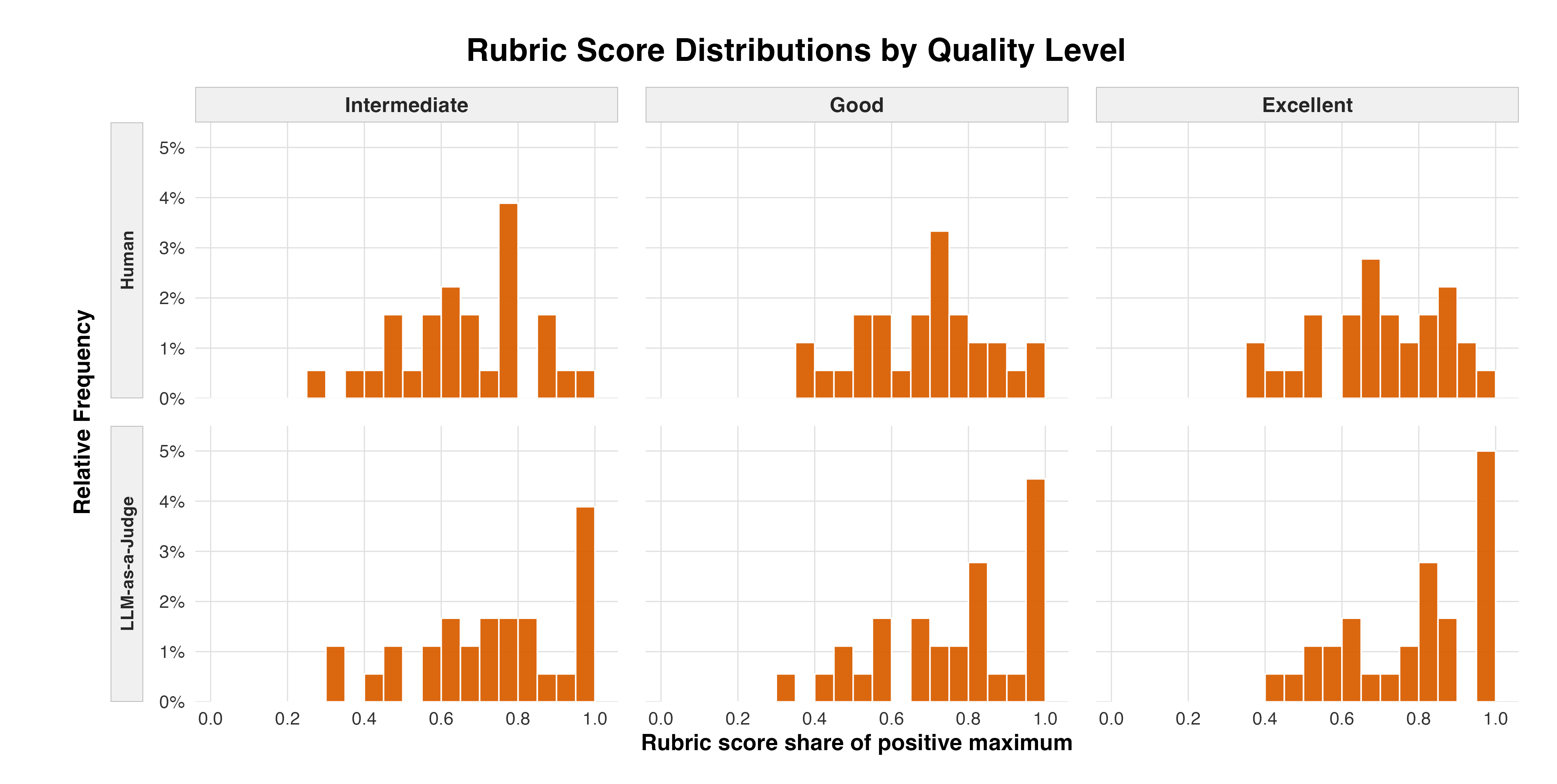}
  \caption{Score distributions by evaluation method and annotator type, aggregated to the quality level for each task. Each panel reports rubric scores as a share of total positive points achievable. Columns separate quality levels; rows separate human evaluators and autograders.}
  \label{fig:quality-level-positive-point-score-distributions}
\end{figure}

\clearpage

\renewcommand{\thetable}{\arabic{table}}
\begin{table}[!t]
  \caption{Adjacent-pair recovery accuracy. For each task, each method's implied score (mean rubric total or Bradley-Terry latent utility) is compared between the two indicated quality levels; a task is counted as correct if the higher-quality level receives the higher score. Cells report the share of tasks correctly ordered. Column headers abbreviate the three quality levels intermediate (I), good (G), and excellent (E). \textit{Adjacent} pools I-vs-G and G-vs-E; \textit{Non-adj.} is I-vs-E.}
  \label{tab:adjacent-pair-accuracy}
  \centering
  \small
  \setlength{\tabcolsep}{4pt}
  \begin{tabular}{llccccc}
    \toprule
    Evaluator & Method & I vs G & G vs E & I vs E & Adjacent & Non-adj. \\
    \midrule
    Human & Rubrics & 0.600 & 0.533 & 0.567 & 0.567 & 0.567 \\
    Human & CJ & 0.933 & 0.833 & 0.967 & 0.883 & 0.967 \\
    \addlinespace[0.35em]
    Autograder & Rubrics & 0.500 & 0.233 & 0.367 & 0.367 & 0.367 \\
    Autograder & CJ & 0.867 & 0.567 & 0.700 & 0.717 & 0.700 \\
    \bottomrule
  \end{tabular}
\end{table}

\clearpage

\subsection{Additional Analyses}
\label{app:additional-analyses} 

\paragraph{Stability across annotators.} We assess comparative-judgment stability within 90 task--quality-pair strata, each containing 11--23 annotations of different output instantiations. Raw dyadic agreement, the probability that two annotations in a stratum both favor the higher-quality or both favor the lower-quality output, is 55.3\%. Leave-one-out consensus agreement is 66.2\% (annotator-clustered bootstrap 95\% CI: 61.8--70.7\%). Higher-quality outputs are preferred in 66.9\% of comparisons (95\% CI: 62.2--71.5\% under the same bootstrap), with leave-one-annotator-out estimates ranging from 66.1\% to 67.5\%. For each annotator and task, we count each output's wins across its two comparisons. Interval Krippendorff's alpha for these counts, treating task--quality levels as units, is 0.133. This is not item-level reliability: annotators evaluated different outputs, and chance correction uses the marginal distribution of win totals that include the overall constructed-quality signal. It therefore should not be directly compared with the rubric IRR calculated on overlapping items.

\paragraph{Surface features and preferences.}
We predicted pairwise preferences from differences in word count, words per sentence, characters per word, Flesch reading ease, Flesch--Kincaid grade level, and counts of list items and headings. Models were fitted separately for human annotators, GPT-5.4, and GPT-5.4-mini using five-fold cross-validation grouped by task. Held-out AUC was 0.67, 0.57, and 0.75, respectively, compared with 0.50 for chance discrimination. Surface features thus predict preferences to some extent, particularly for GPT-5.4-mini.

\paragraph{Criterion-level rubric discrimination.} For each task, we compared the human scores on individual criteria that award positive points. We compared every scored output at one quality level with every scored output at another quality level, checking whether the higher-quality output received more, equal, or fewer points on the same criterion. A tie means that both outputs received the same points on that criterion, such as zero points each, one point each, or equal partial scores. Approximately three-quarters of these comparisons were ties. When scores differed, the higher-quality output more often received more points (Table~\ref{tab:criterion-discrimination}).

We separately examined the effect of negative deductions. We removed these deductions, averaged the resulting scores within each task and quality level, and checked whether the higher-quality level had the higher average score. Across the 30 tasks, removing deductions increased correct ordering from 60\% to 63\% for good versus intermediate, left it unchanged at 53\% for excellent versus good, and increased it from 57\% to 63\% for excellent versus intermediate.

\begin{table}[ht]
\centering
\small
\caption{Criterion-level discrimination for positive rubric criteria scored by human annotators. Columns report the share of comparisons in which the higher-quality output receives more, equal, or fewer points.}
\label{tab:criterion-discrimination}
\begin{tabular}{lrrr}
\toprule
Comparison & More & Equal & Fewer \\
\midrule
Excellent vs.\ Good & 13.3\% & 75.7\% & 11.1\% \\
Good vs.\ Intermediate & 13.1\% & 75.2\% & 11.7\% \\
Excellent vs.\ Intermediate & 14.1\% & 75.4\% & 10.5\% \\
\bottomrule
\end{tabular}
\end{table} 

\paragraph{Rubric alignment.} For five randomly sampled rubrics comprising 84 items, Harvey supplied open-text explanations of each item's purpose. We separately used GPT-5.4 to code alignment with the six quality dimensions and manually validated the coding. Of the 84 items, 67 (80\%) map directly to these dimensions, predominantly completeness. The remaining items generally concern formatting, style, or error penalties. This overlap does not establish equal coverage of the six dimensions. 

\paragraph{Expanded autograder evaluation.} Table~\ref{tab:expanded-autograders} reports the per-judgment win-rate comparison across eight autograders. Every model exhibits an advantage for comparative judgments.  

\begin{table}[ht]
\centering
\small
\caption{Expanded autograder results. $\bar{W}^{\mathrm{CJ}}$ and $\bar{W}^{\mathrm{rub}}$ are task-averaged per-judgment win rates, with rubric ties counted as 0.5; $\widehat{D}_{W}$ is their difference.}
\label{tab:expanded-autograders}
\begin{tabular}{lrrr}
\toprule
Evaluator & $\bar{W}^{\mathrm{CJ}}$ & $\bar{W}^{\mathrm{rub}}$ & $\widehat{D}_{W}$ \\
\midrule
Human & 0.669 & 0.542 & 0.127 \\
GPT-5.4 & 0.599 & 0.427 & 0.172 \\
GPT-5.4-mini & 0.702 & 0.449 & 0.254 \\
Claude Opus 4.8 & 0.890 & 0.465 & 0.425 \\
Claude Sonnet 5 & 0.852 & 0.446 & 0.406 \\
Gemini 3.1 Pro & 0.820 & 0.533 & 0.287 \\
Gemini 3.6 Flash & 0.819 & 0.560 & 0.259 \\
Qwen 3.7 Max & 0.841 & 0.524 & 0.317 \\
GLM-5.2 & 0.875 & 0.484 & 0.390 \\
\bottomrule
\end{tabular}
\end{table}

\paragraph{Autograder deductions.}
For GPT-5.4, mean positive rubric points increase from 9.4 to 9.9 to 10.3 across intermediate, good, and excellent outputs, while mean negative points become more negative, from $-1.9$ to $-2.3$ to $-3.0$. Removing deductions raises task-level ordering recovery from 0.50 / 0.23 / 0.37 to 0.73 / 0.57 / 0.83 for intermediate--good, good--excellent, and intermediate--excellent comparisons, respectively. These are shares of correctly ordered task-level score pairs, as in Table~\ref{tab:adjacent-pair-accuracy}, rather than per-judgment win rates. For the original rubric results in that table, exact binomial 95\% confidence intervals include 0.5 in every quality-pair cell except autograder good--excellent; that departure is also nonsignificant after Bonferroni correction.  

\paragraph{Mixed quality profiles.} We assign intermediate, good, and excellent dimension levels numerical values of 1, 2, and 3, respectively, and generate outputs using all six-dimensional profiles averaging 1.5 or 2.5. Their one-point separation matches the adjacent comparisons in the original construction. Using GPT-5.4, we evaluate 1,500 pairs across the 30 tasks under both methods. On these mixed profiles, per-judgment win rates are 0.58 for comparative judgments and 0.45 for rubrics, a difference of 0.13. The difference relative to the method advantage on adjacent original profiles is $-0.03$ (95\% CI: $[-0.09, 0.03]$, $p=0.36$). We estimate its standard error using 2,000 paired bootstrap resamples of tasks, preserving the pairing across methods and profile constructions, and compute a normal-approximation confidence interval and two-sided $p$-value. We therefore detect no change in the method difference in this experiment, without interpreting the nonsignificant result as evidence of equivalence.  

\subsection{Ethical Considerations}
\label{app:ethical-considerations}

\paragraph{Access to justice.}
Our evaluation setting focuses on economically valuable legal work, which is the arena in which modern legal AI tools are being developed and deployed. However, this should not be read as a claim that the evaluation of legal AI is most pressing for these particular applications. In other settings such as legal aid and public defense, the development of legal AI tools around client needs demands a similarly rigorous approach to evaluation. Future work should explore how the notion of quality is operationalized in these settings, paying particular attention to the ethical considerations that arise. 

\paragraph{Safeguards for responsible release.}
Because the benchmark concerns legal work products, we apply safeguards to the released materials. The released benchmark materials are intended for research and evaluation, not for providing legal advice. Before release, tasks, outputs, and annotations were screened for confidential client information, privileged material, and personally identifying information. Released materials are accompanied by documentation describing their intended research use and limitations.

\subsection{Assets and Licenses}
\label{app:assets-and-licenses}

\paragraph{Existing assets.}
The 30 base tasks are drawn from Harvey's BigLaw Bench and used with permission from the original rights holder. 

\paragraph{Released assets.}
The authors release the \textsc{JudgmentBench} annotation-level dataset introduced in this work, including the rubric scores, pairwise preference judgments, annotation metadata, analyses, and derived evaluation files, under the MIT License.

\clearpage
\section{Model Prompts}
\label{app:model-prompts}

This appendix reports the prompts used to generate synthetic work products and to conduct LLM-based validation and evaluation analyses. Prompt text is shown with placeholders where task-specific inputs, quality-level variables, or analysis controls are inserted.

\subsection{Prompt for Generating Quality-Controlled Work Products}
\label{app:prompt-generation}

We used Claude Opus 4.6 to generate the work products shown to annotators and used in the downstream evaluation pipelines. The same prompt template was used for all three quality levels; only the quality-profile variable differed across the \textit{intermediate}, \textit{good}, and \textit{excellent} conditions. At each quality level, all six values under \texttt{"quality\_profile"} are set to the same value corresponding to that quality level (e.g., for the \textit{excellent} response, \texttt{"analytical\_depth"}, \texttt{"precision"}, \texttt{"completeness"}, and so on are all set to \texttt{"excellent"}).

\prompttcbinput{prompts/quality_controlled_generation_prompt.txt}
  {Prompt for Quality-Controlled Work Product Generation}
  {0}
  {quality-controlled-generation-prompt}

\subsection{Prompt for Validating Quality-Level Separation}
\label{app:prompt-quality-validation}

We used GPT-5.4 as an LLM-as-a-Judge to validate that adjacent constructed quality levels were sufficiently distinguishable in within-task pairwise comparisons.

\prompttcbinput{prompts/quality_level_validation_prompt.txt}
  {Prompt for Validating Quality-Level Separation}
  {0}
  {quality-level-validation-prompt}
\clearpage
\subsection{Prompt for Autograder Rubric-Based Scoring and Comparative Judgment}
\label{app:prompt-llm-judge-analysis}

We used the models listed in Table~\ref{tab:expanded-autograders} to conduct autograder rubric-based scoring and comparative-judgment evaluation on the dataset. The same prompt template was used across analysis modes, with the evaluation protocol controlled by a variable.

\prompttcbinput{prompts/llm_judge_rubric_prompt.txt}
  {Prompt for Autograder Rubric-Based Scoring}
  {0}
  {llm-judge-rubric-prompt}
\clearpage
\prompttcbinput{prompts/llm_judge_pref_prompt.txt}
  {Prompt for Autograder Comparative Judgment}
  {0}
  {llm-judge-pref-prompt}

\clearpage

\section{Expert Annotator Instructions and Human Subjects Details}
\label{app:human-subjects}

\subsection{Participant Instructions}
\begin{figure}[p]
  \centering
  \includegraphics[width=\textwidth]{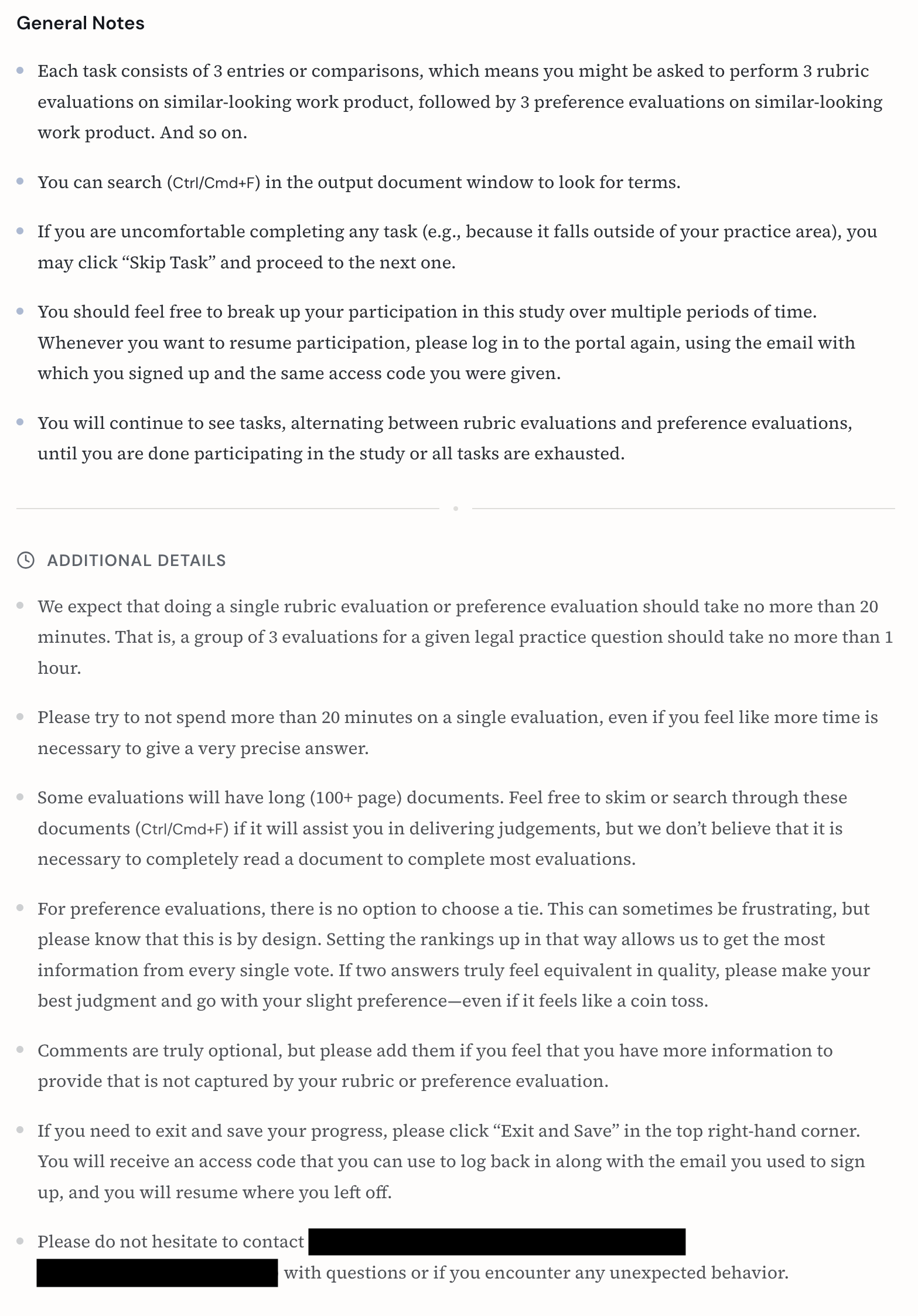}
  \caption{General explanatory notes that annotators viewed in the interface before proceeding to evaluations.}
  \label{fig:guide}
\end{figure}

\begin{figure}[p]
  \centering
  \includegraphics[width=\textwidth]{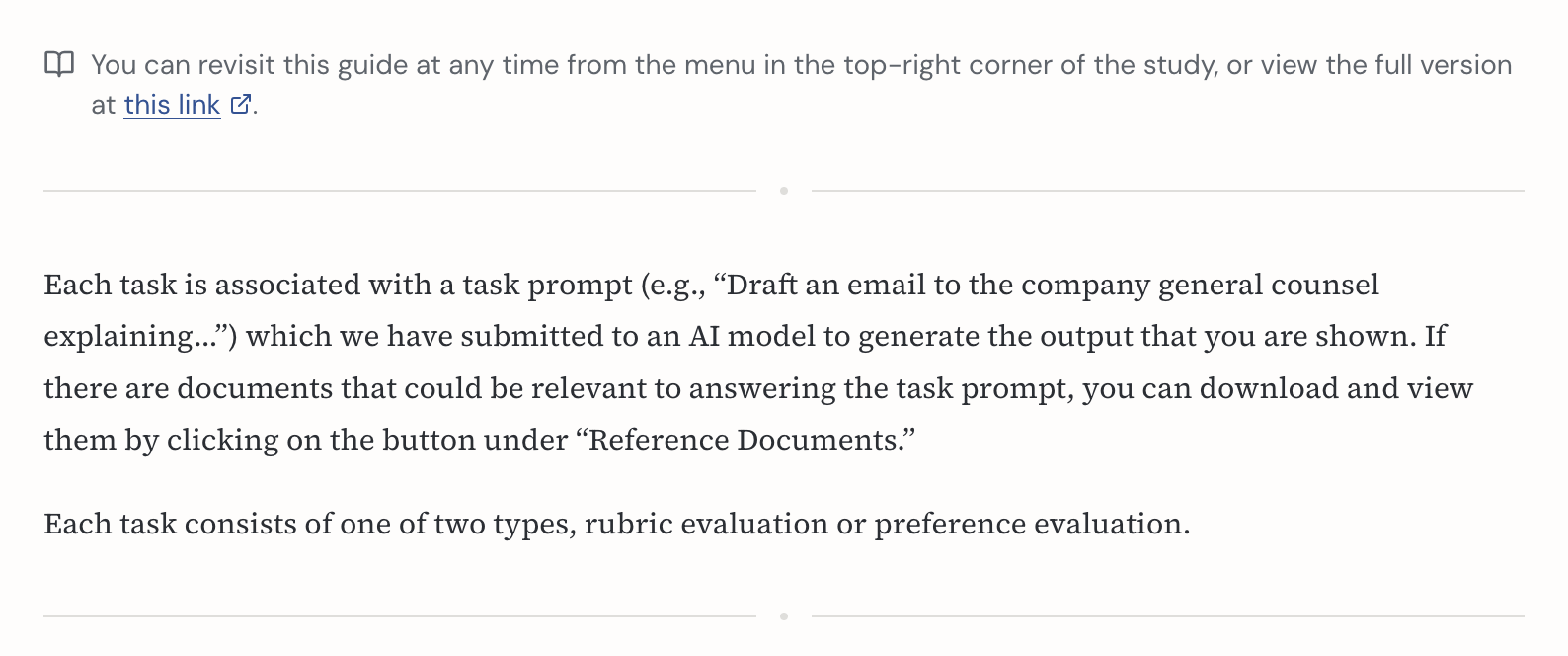}
  \caption{Explanation of tasks, prompts, and documents that annotators viewed in the interface before proceeding to evaluations.}
  \label{fig:task-types}
\end{figure}

Appendix Figures~\ref{fig:guide} and ~\ref{fig:task-types} show general instructions presented to annotators.

\subsection{Preference and Rubric Instructions}
\begin{figure}[p]
  \centering
  \includegraphics[width=\textwidth]{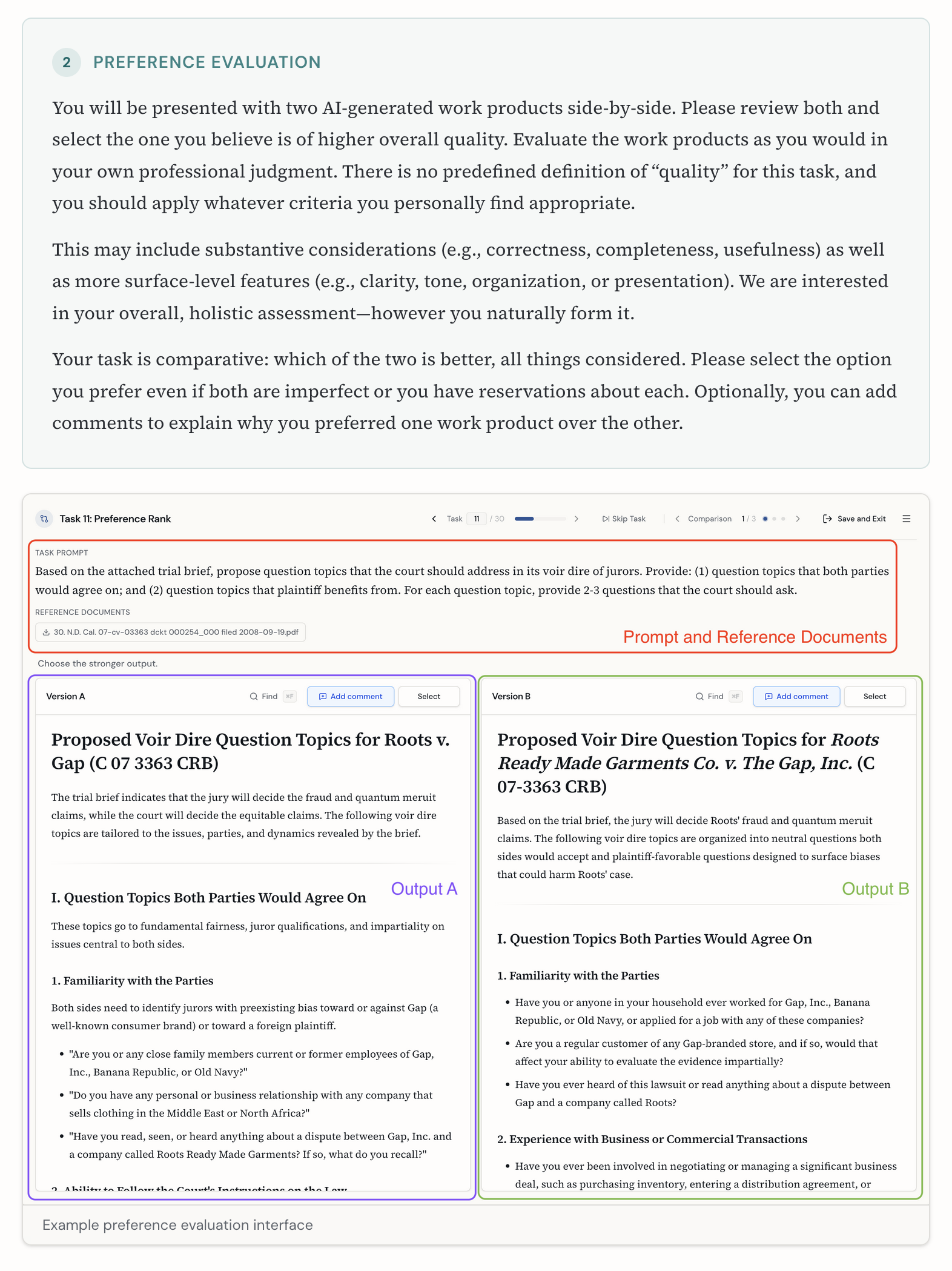}
  \caption{Explanation of comparative judgments with an interface example.}
  \label{fig:pref-instructions}
\end{figure}

\begin{figure}[p]
  \centering
  \includegraphics[width=\textwidth]{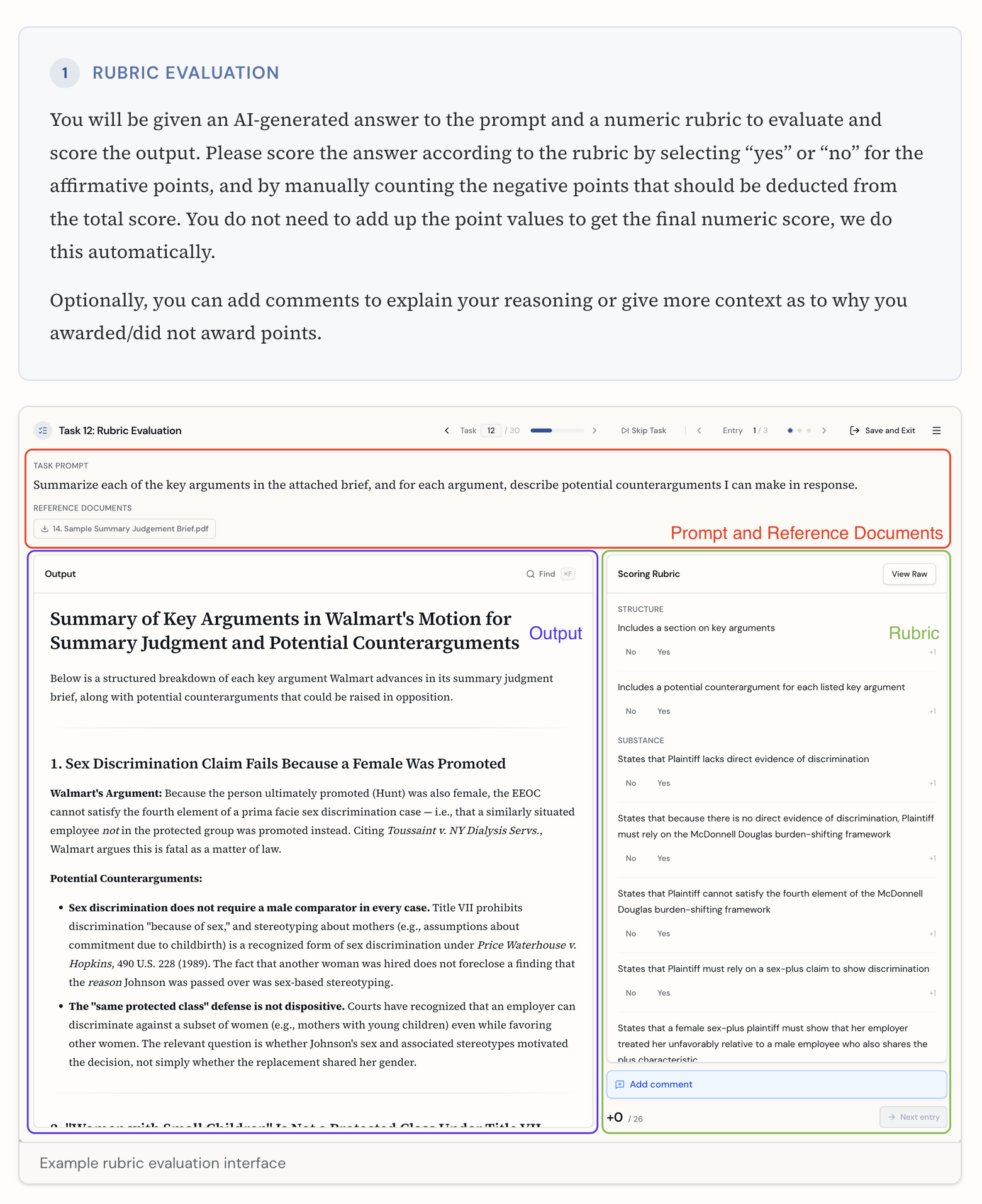}
  \caption{Explanation of rubric evaluations with an interface example.}
  \label{fig:rubric-instructions}
\end{figure}

Appendix Figures~\ref{fig:pref-instructions} and~\ref{fig:rubric-instructions}
show examples of the interfaces used in comparative-judgment and rubric evaluations.

\subsection{Ethics Review}
\label{app:ethics-review}
This study was reviewed by the authors' institutional review board and determined to be exempt human-subjects research under protocol ID 86215. The consent materials disclosed the study purpose, expected time commitment, compensation, data handling practices, and reasonably foreseeable risks.

\subsection{Compensation}
Annotators from law firms participated through the pro bono practices of their firms and did not receive additional compensation. All annotators recruited through Snorkel AI were paid by Snorkel AI, at rates typically ranging from \$100 to \$150 per hour, depending on seniority and practice area.

\clearpage

\clearpage

\clearpage

\clearpage

\newpage
\section*{NeurIPS Paper Checklist}

\begin{enumerate}

\item {\bf Claims}
    \item[] Question: Do the main claims made in the abstract and introduction accurately reflect the paper's contributions and scope?
    \item[] Answer: \answerYes{} %
    \item[] Justification: The claims made in the abstract and introduction accurately reflect the paper's contributions and scope.

\item {\bf Limitations}
    \item[] Question: Does the paper discuss the limitations of the work performed by the authors?
    \item[] Answer: \answerYes{} %
    \item[] Justification: The limitations are discussed in Section~\ref{app:limitations}.

\item {\bf Theory assumptions and proofs}
    \item[] Question: For each theoretical result, does the paper provide the full set of assumptions and a complete (and correct) proof?
    \item[] Answer: \answerNA{} %
    \item[] Justification: There are no theoretical results.

    \item {\bf Experimental result reproducibility}
    \item[] Question: Does the paper fully disclose all the information needed to reproduce the main experimental results of the paper to the extent that it affects the main claims and/or conclusions of the paper (regardless of whether the code and data are provided or not)?
    \item[] Answer: \answerYes{} %
    \item[] Justification: The released study dataset and analysis code are made fully available. Section~\ref{sec:dataset-construction} describes the dataset construction, additional methodological details are in Appendix~\ref{app:detailed-methodology}, and the prompts are in Appendix~\ref{app:model-prompts}.  

\item {\bf Open access to data and code}
    \item[] Question: Does the paper provide open access to the data and code, with sufficient instructions to faithfully reproduce the main experimental results, as described in supplemental material?
    \item[] Answer: \answerYes{} %
    \item[] Justification: Open access is provided, and the repository contains instructions to faithfully replicate the main experimental results.

\item {\bf Experimental setting/details}
    \item[] Question: Does the paper specify all the training and test details (e.g., data splits, hyperparameters, how they were chosen, type of optimizer) necessary to understand the results?
    \item[] Answer: \answerYes{} %
    \item[] Justification: The experimental setting is described in Section~\ref{sec:dataset-construction}, including the dataset construction, controlled quality levels, expert annotator sample, and annotation protocols. Additional methodological details, including aggregation, outcome metrics, timing analysis, and bootstrap procedures, are provided in Appendix~\ref{app:detailed-methodology}.

\item {\bf Experiment statistical significance}
    \item[] Question: Does the paper report error bars suitably and correctly defined or other appropriate information about the statistical significance of the experiments?
    \item[] Answer: \answerYes{} %
    \item[] Justification: Uncertainty estimates are reported for the main result using a hierarchical bootstrap procedure described in Appendix~\ref{app:detailed-methodology}.

\item {\bf Experiments compute resources}
    \item[] Question: For each experiment, does the paper provide sufficient information on the computer resources (type of compute workers, memory, time of execution) needed to reproduce the experiments?
    \item[] Answer: \answerYes{} %
    \item[] Justification: The experiments do not involve model training or fine-tuning. Compute requirements are limited to API-based generation and evaluation and lightweight statistical analysis, including Bradley--Terry aggregation and hierarchical bootstrap estimation, which can be run on a standard CPU workstation.
    
\item {\bf Code of ethics}
    \item[] Question: Does the research conducted in the paper conform, in every respect, with the NeurIPS Code of Ethics \url{https://neurips.cc/public/EthicsGuidelines}?
    \item[] Answer: \answerYes{} %
    \item[] Justification: The paper conforms, in every respect, with the NeurIPS Code of Ethics.

\item {\bf Broader impacts}
    \item[] Question: Does the paper discuss both potential positive societal impacts and negative societal impacts of the work performed?
    \item[] Answer: \answerYes{} %
    \item[] Justification: Societal impacts are discussed in Section~\ref{sec:discussion} and Appendix~\ref{app:ethical-considerations}.
    
\item {\bf Safeguards}
    \item[] Question: Does the paper describe safeguards that have been put in place for responsible release of data or models that have a high risk for misuse (e.g., pre-trained language models, image generators, or scraped datasets)?
    \item[] Answer: \answerYes{} %
    \item[] Justification: The safeguards are described in ~\ref{app:ethical-considerations}.

\item {\bf Licenses for existing assets}
    \item[] Question: Are the creators or original owners of assets (e.g., code, data, models), used in the paper, properly credited and are the license and terms of use explicitly mentioned and properly respected?
    \item[] Answer: \answerYes{} %
    \item[] Justification: Permission and licensing details for the benchmark materials are discussed in Appendix~\ref{app:assets-and-licenses}.

\item {\bf New assets}
    \item[] Question: Are new assets introduced in the paper well documented and is the documentation provided alongside the assets?
    \item[] Answer: \answerYes{} %
    \item[] Justification: All assets are well documented.

\item {\bf Crowdsourcing and research with human subjects}
    \item[] Question: For crowdsourcing experiments and research with human subjects, does the paper include the full text of instructions given to participants and screenshots, if applicable, as well as details about compensation (if any)? 
    \item[] Answer: \answerYes{} %
    \item[] Justification: Annotator instructions, annotation-interface information, and compensation details are provided in Appendix~\ref{app:human-subjects}.

\item {\bf Institutional review board (IRB) approvals or equivalent for research with human subjects}
    \item[] Question: Does the paper describe potential risks incurred by study participants, whether such risks were disclosed to the subjects, and whether Institutional Review Board (IRB) approvals (or an equivalent approval/review based on the requirements of your country or institution) were obtained?
    \item[] Answer: \answerYes{} %
    \item[] Justification: Details of risks and IRB exemption are explained in Section ~\ref{app:ethics-review}.

\item {\bf Declaration of LLM usage}
    \item[] Question: Does the paper describe the usage of LLMs if it is an important, original, or non-standard component of the core methods in this research? Note that if the LLM is used only for writing, editing, or formatting purposes and does \emph{not} impact the core methodology, scientific rigor, or originality of the research, declaration is not required.
    \item[] Answer: \answerYes{} %
    \item[] Justification: The usage of LLMs is described as it pertains to this research.

\end{enumerate}

\end{document}